# BEDI: A Comprehensive Benchmark for Evaluating Embodied Agents on UAVs

**Mingning Guo[1,#], Mengwei Wu[1,#], Jiarun He[1], Shaoxian Li[1], Haifeng Li[1], Chao Tao[1,*]**

1 School of Geosciences and Info-Physics, Central South University, Changsha 410083, China; 225007012@csu.edu.cn, 235007016@csu.edu.cn, 245011029@csu.edu.cn, lishaoxian@csu.edu.cn, lihaifeng@csu.edu.cn, kingtaochao@csu.edu.cn.

\# These authors contributed equally to this work.

* Correspondence author: kingtaochao@csu.edu.cn.

**Abstract**: With the rapid advancement of low-altitude remote sensing and Vision-Language Models (VLMs), Embodied Agents based on Unmanned Aerial Vehicles (UAVs) have shown significant potential in autonomous tasks. However, current evaluation methods for UAV-Embodied Agents (UAV-EAs) remain constrained by the lack of standardized benchmarks, diverse testing scenarios and open system interfaces. To address these challenges, we propose BEDI (Benchmark for Embodied Drone Intelligence), a systematic and standardized benchmark designed for evaluating UAV-EAs. Specifically, we introduce a novel Dynamic Chain-of-Embodied-Task paradigm based on the perception-decision-action loop, which decomposes complex UAV tasks into standardized, measurable subtasks. Building on this paradigm, we design a unified evaluation framework encompassing six core sub-skills: semantic perception, spatial perception, motion control, tool utilization, task planning and action generation. Furthermore, we develop a hybrid testing platform that incorporates a wide range of both virtual and real-world scenarios, enabling a comprehensive evaluation of UAV-EAs across diverse contexts. The platform also offers open and standardized interfaces, allowing researchers to customize tasks and extend scenarios, thereby enhancing flexibility and scalability in the evaluation process. Finally, through empirical evaluations of several state-of-the-art (SOTA) VLMs, we reveal their limitations in embodied UAV tasks, underscoring the critical role of the BEDI benchmark in advancing embodied intelligence research and model optimization. By filling the gap in systematic and standardized evaluation within this field, BEDI facilitates objective model comparison and lays a robust foundation for future development in this field. Our benchmark is now publicly available at https://github.com/lostwolves/BEDI.

## 1. Introduction

Unmanned aerial vehicles (UAVs) are widely used in applications such as disaster relief (Khan et al., 2022), environmental monitoring (Qin et al., 2024), agricultural surveillance (Deng et al., 2018), field patrolling (Giuseppi et al., 2021), and infrastructure inspection (Lekidis et al., 2022), owing to their efficiency, flexibility, mobility, and low operational costs. However, most UAVs still depend heavily on remote human control, which becomes inefficient in complex and dynamic environments. As task complexity increases and the demand for autonomy grows, there is a pressing need to develop UAV-Embodied Agents (UAV-EAs) — autonomous systems capable of independently performing environmental perception, task decision-making, and execution (Duan et al., 2022; Levine et al., 2018; Tian et al., 2025).

To address the challenges of operating in dynamic and open scenarios, UAV-EAs must transcend the limitations of traditional rule-based control systems. These systems often struggle with real-time adaptation to environmental uncertainties and complex task requirements. The core capabilities required for robust performance in such contexts encompass three key dimensions: **environmental perception, spatial reasoning and autonomous decision-making** (Campos-Macías et al., 2021). A systematic benchmarking framework is therefore critical for quantitatively evaluating these interdependent competencies, particularly in dynamic and open scenarios. However, current benchmarks fail to accurately assess the capabilities of UAV-EAs due to three key limitations. Firstly, evaluation metrics and methodologies differ significantly across studies, lacking a standardized framework. Secondly, most benchmarks are based on predefined, static scenarios, overlooking the challenges of dynamic environments. Thirdly, evaluation platforms are often closed systems, limiting compatibility with externally developed UAV-EAs and hindering reproducible testing. To address these limitations, we propose **BEDI (Benchmark for Embodied Drone Intelligence)**, a benchmark specifically designed for evaluating UAV-EAs.

**(1) Standardized Task Evaluation Framework:** A standardized evaluation framework is crucial for objectively assessing UAV-EA capabilities, but existing benchmarks often lack consistency and rely on task-specific metrics, complicating cross-comparison. For instance, CityNav (Lee et al., 2024) and AerialVLN (Liu et al., 2023) both focus on navigation but use varying combinations of metrics such as Success Rate (SR), Oracle Success Rate (OSR), Navigation Error (NE), and other weighted success measures. In decision-oriented tasks, benchmarks like EmbodiedCity (Gao et al., 2024) rely on traditional language metrics (e.g., BLEU, ROUGE), while AeroVerse (Yao et al., 2024) introduces metrics based on Large Language Models (LLMs) such as SCENE and PLAN. To address this issue, we propose the **Dynamic Chain-of-Embodied-Task,** a unifying paradigm that reformulates diverse embodied tasks into a standardized sequential structure: *"Perception*$_1$ → *Decision*$_1$ → *Action*$_1$ → *Perception*$_2$ → *Decision*$_2$ → *Action*$_2$ *..."*, as illustrated in **Fig. 1**. Within this framework, the UAV operates through a closed **perception–decision–action loop**: it acquires environmental data via onboard sensors (*perception*), processes the information to formulate decisions such as path planning (*decision*), and executes corresponding control commands (*action*) to interact with the environment. These interactions subsequently influence new perceptual inputs, forming a continuous feedback loop. Building upon this, we develop a standardized evaluation framework comprising step-level, loop-level, and task-level metrics that systematically assess six core sub-skills within the perception-decision-action loop: semantic perception, spatial perception, motion control, tool utilization, task planning and action generation. This hierarchical framework enables comprehensive and standardized assessment of UAV-EAs, ranging from individual capability evaluation to complete task performance measurement, effectively addressing the ambiguity and inconsistency limitations in existing benchmarks.

**(2) Comprehensive Testing Scenarios Across Real-Virtual and Dynamic-Static Dimensions:** UAV-EAs operate in dynamic, diverse real-world environments, requiring testing scenarios that reflect task complexity and environmental variability. However, most existing benchmarks rely on simplified virtual environments or one-turn evaluations using static images. For example, OmniDrones (Xu et al., 2024) evaluates basic UAV control tasks such as hovering and obstacle avoidance in empty scenes, while AeroVerse and EmbodiedCity assess UAV-EAs using high-fidelity simulations and single-view urban scenes, respectively. These approaches fail to capture real-world complexity and overlook the interdependencies between the perception, decision, and action stages in the dynamic task loop. In contrast, as illustrated in **Fig. 1**, we develop a hybrid testing platform that incorporates a wide range of both virtual and real-world scenarios to create the Benchmark for Embodied Drone Intelligence: BEDI. Specifically, in the static real-world setting, we collect real UAV imagery under diverse conditions, including various terrains (e.g., ocean, city, hills), scenarios (e.g., fire, flood, traffic), and lighting conditions (e.g., bright, dim, dark), to evaluate the stability and generalizability of UAV-EAs across real-world contexts. In the dynamic real-world setting, we record UAV flight videos and extract key frames from single-task execution sequences. We then employ a dynamic interactive testing approach to evaluate the stability and reliability of UAV agents in performing dynamic tasks. In the dynamic virtual setting, we simulate complex tasks like cargo delivery, building firefighting, and moving target tracking. These tasks evaluate UAV-EAs' perception, decision, and action abilities under dynamic conditions. By synthesizing real-world and simulated environments with both static and dynamic conditions, BEDI provides a more comprehensive evaluation of the core capabilities of UAV-EAs compared with existing benchmarks.

**(3) Open Interfaces for UAV-EA Task Evaluation:** Open evaluation interfaces are essential for enabling researchers to assess UAV-EA performance. Although benchmarks such as AerialVLN, AeroVerse, and EmbodiedCity provide high-fidelity physical environments to evaluate UAV-EAs, most of them remain closed-source. The lack of openness hinders fair evaluation and makes it difficult to reveal the limitations of current UAV-EAs. To address this, BEDI publicly releases real-world test samples as accessible datasets and provides a modular testing interface for virtual environments. Specifically, we offer a detailed urban simulation built with Unreal Engine and design a user-friendly, open task interface. This interface allows for seamless integration of custom UAV-EAs and efficient scenario expansion tailored to research requirements. Through its open architecture, BEDI overcomes the constraints of existing benchmarks, offering the community a framework renowned for its accessibility, flexibility, and extensibility.

Through evaluations on the BEDI benchmark, we assessed several state-of-the-art VLMs, such as the GPT series, Llama-3.2-Vision-Instruct (Meta AI, 2024), LLaVA-OneVision (Li et al., 2024), and Qwen2-VL (P. Wang et al., 2024), which collectively revealed substantial limitations in their ability to handle embodied UAV tasks. As the first unified and openly accessible benchmark for UAV-EAs, BEDI establishes a critical foundation for future research in embodied intelligence. Its main contributions are as follows:

- We establish a standardized hierarchical evaluation framework based on the Dynamic Chain-of-Embodied-Task paradigm. In contrast to existing benchmarks that rely on fragmented and inconsistent metrics, our framework systematically assesses UAV-EAs through step-level, loop-level, and task-level metrics, enabling comprehensive quantification of three core capabilities—perception, decision, and action—along with their six underlying sub-skills. This structured approach offers a unified and reproducible standard for evaluating agent performance across diverse embodied tasks.
- We introduce BEDI, the first fully open benchmark for UAV embodied agents, integrating a wide spectrum of testing scenarios across real–virtual and static–dynamic dimensions. By publicly releasing real-world image and video datasets and providing flexible virtual task interfaces built on Unreal Engine, BEDI supports scalable and reproducible evaluation under varied environmental conditions. This open and extensible design addresses the limitations of closed benchmarks and fosters transparent, comparable, and community-driven research in UAV embodied intelligence.

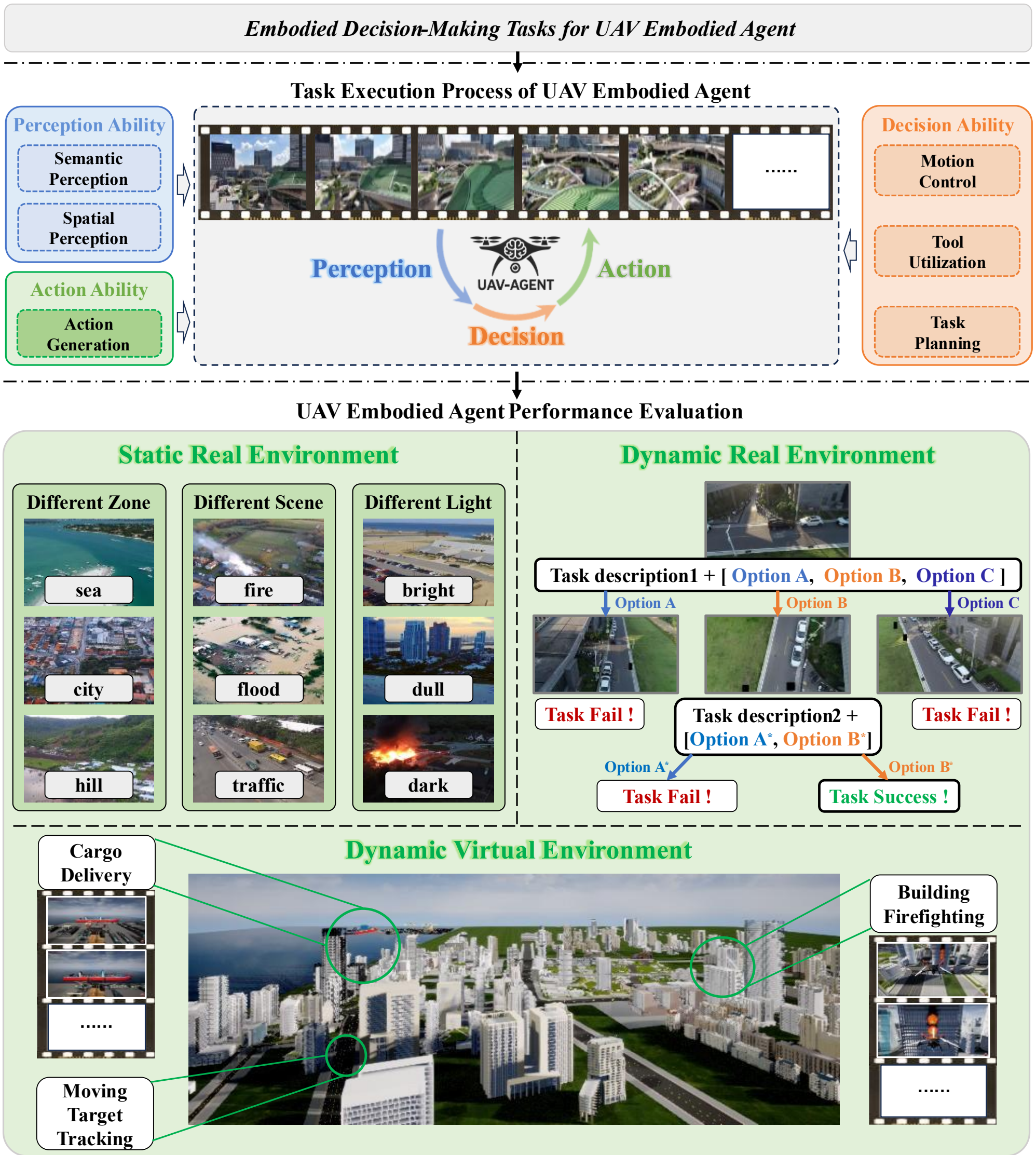

Fig. 1 Overview of our proposed BEDI framework. The task execution process of a UAV-EA is standardized into a cyclic interaction loop, primarily including the agent's perception, decision, and action stages. For task evaluation environments, BEDI includes a static real environment based on actual UAV imagery, a dynamic real environment based on actual UAV video and a dynamic virtual environment designed with multiple embodied tasks.

## 2. The Core Abilities of UAV Embodied Agent

### 2.1 The Dynamic Chain-of-Embodied-Task for UAV Embodied Agent

Human interaction with the physical world is fundamentally driven by a continuous feedback loop: we perceive environments and make decisions based on sensory input, then act upon the world, which in turn generates new perceptual data. This cycle fosters adaptive behavior through constant refinement. For instance, when walking out of a maze, a person integrates sensory input to estimate direction, continuously adjusts movement based on feedback, and draws upon prior experience to make decisions—ultimately finding the way out. Similarly, UAV-EAs could follow a comparable loop in task execution. For example, in a ship-type recognition task, the UAV uses onboard sensors to capture visual data, extracting features like hull identification numbers (Guo et al., 2025). If the view is obscured due to angle, distance, or occlusion, the UAV-EA adapts by adjusting its behavior, such as moving closer, zooming in, or switching sensors, to capture clearer images. This updated perception then feeds into the next decision cycle, continuing until the task is completed.

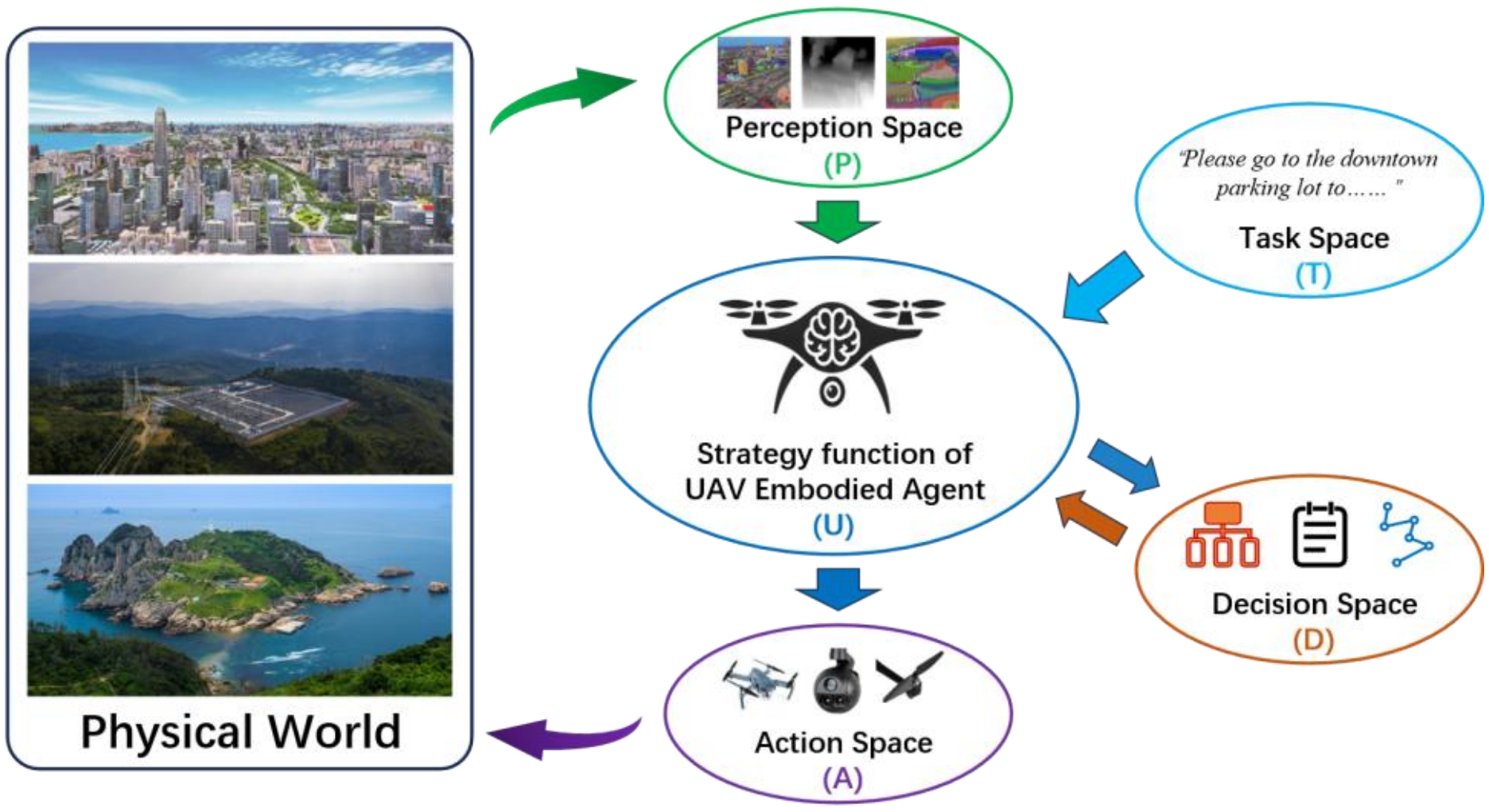

Fig. 2 Interaction loop of a UAV-EA during task execution.

Based on the above observation, we propose the Dynamic Chain-of-Embodied-Task paradigm, which mirrors the human-like perception-decision-action loop. In this paradigm, an agent performs an alternating sequence of perception, decision, and action, such as "$Perception_1 \rightarrow Decision_1 \rightarrow Action_1 \rightarrow Perception_2 \rightarrow Decision_2 \rightarrow Action_2$ …", continuously refining its behavior in dynamic and uncertain environments. As shown in **Fig. 2**, under this paradigm, the task execution process of a UAV-EA can be modeled as a Markov Decision Process (MDP) defined as $M = (U (T, P, D), A)$, where:

- **T denotes the task description space**, which typically encodes task objectives in natural language. This facilitates the UAV-EA's understanding and planning of task goals.
- **P represents the perception space**, encompassing semantic scene information and spatial relationships among observed objects. It captures detailed characteristics such as semantic labels, geometric features, texture attributes, and the relative spatial configuration between different targets.
- **D is the decision space**, containing the set of possible decisions made by the agent based on task goals and perceptual inputs. It involves sub-goal decomposition, priority ranking, resource allocation, and the selection of execution strategies that guide agent behavior.
- **A denotes the action space**, which includes the executable actions of the UAV platform, such as physical movement, camera viewpoint adjustment, and sensor zoom control.
- **U is the strategy function of the UAV-EA during task execution**. It integrates information from T, P, and D to generate appropriate control commands in A.

## 2.2 Core Capability 1: Perception

**Perception** is a fundamental mechanism through which humans understand and interact with the world. Studies show that about 80% of the information humans receive comes from visual perception, with nearly one-third of the cerebral cortex involved in processing it (Stark, 2001). Biologists hypothesize a critical link between multimodal perception and behavioral development (deCharms and Zador, 2000), and neuroscientists have provided evidence supporting a strong connection between perceptual ability and biological activity (Held and Hein, 1963). These findings highlight the key role of visual perception in human learning and evolution (De Sousa et al., 2022).

For UAV-EAs, perception in three-dimensional environments enables the agent to extract task-relevant information from the physical world into the perceptual space (P). Specifically, 3D perception involves two core capabilities: **semantic perception** and **spatial perception**. Semantic perception refers to the UAV-EA's capacity to identify and classify objects within the task environment. Spatial perception consists of two components: directional perception, which allows the UAV-EA to recognize the relative orientation of a target, and distance perception, which facilitates the assessment of relative spatial distances and the judgment of object proximity.

### 2.2.1 Semantic Perception

Semantic perception enables UAV-EAs to recognize, interpret, and understand object meanings and functions in task environments. Traditional approaches often employ self-supervised learning strategies to develop powerful encoders for analyzing visual information from onboard sensors (Tao et al., 2023). These models perform well in fundamental tasks like object classification and scene understanding (Berg et al., 2022), particularly under ideal conditions where images are clear and objects are fully visible. However, models trained exclusively on visual supervision exhibit significant limitations in open-world scenarios, as their sole reliance on visual signals compromises perceptual reliability and diminishes their efficacy for fine-grained semantic tasks. To address these challenges, recent methods introduce text-guided learning paradigms that align visual and linguistic representations, enhancing semantic perception in dynamic environments. For instance, the Maria (Liang et al., 2021) uses large-scale image-text pairs and external knowledge to build a robust text-to-image retrieval framework, enabling sensory-contextual grounding for improved semantic understanding. ReSee (Tu et al., 2023) extends this approach to dialogue systems by integrating visual semantics at entity and dialogue-turn levels, improving response coherence in visually grounded conversations. Similarly, TikTalk (Lin et al., 2023) incorporates real-world conversational data and structured knowledge for video-based multimodal dialogue, supporting context-aware interactions.

However, these models are limited by their passive perceptual approach. They perform well when the target is clearly specified by the user (e.g., 'What is object 1 in the image?'), and can provide accurate semantic descriptions. In contrast, their performance declines significantly when locating objects based on function or attributes (e.g., "Which object can transport cargo?") or when distinguishing among multiple similar objects in complex scenes.

### 2.2.2 Spatial Perception

Spatial perception enables the UAV-EA to perceive and reason about object positions, orientations, and their relationships in space. It can be subdivided into two sub-capabilities: directional perception and distance perception.

Directional perception relies on the agent's ability to localize objects and reason about their relative orientation. The agent typically detects both reference and target objects in an image and infers their relative spatial direction. For example, Gkioxari et al. (Gkioxari et al., 2018) developed a dual-stream network that separately localizes humans and surrounding objects, then employs an interaction reasoning module to deduce spatial relationships. Subsequent transformer-based models (e.g., HOTR (Kim et al., 2021)) and diffusion models (e.g., InterDiff (Xu et al., 2023)) build upon prior architectures to improve object localization and enhance directional reasoning capabilities. However, these approaches often struggle with fine-grained directional reasoning, such as clock-face orientation. With the advent of LLMs (Wu et al., 2025), some approaches have integrated language understanding with visual input to enhance spatial reasoning. For example, SpatialRGPT (Cheng et al., 2024) first determines target locations in images and combines visual inputs with language descriptions to leverage LLMs for relative orientation reasoning. LLMI3D (Yang et al., 2025) introduces 3D-aware token decoding for precise geometric regression, improving accuracy in 3D spatial reasoning. However, UAV-EA research typically focuses on spatial reasoning from the agent's first-person perspective, where models designed for third-person contexts often perform poorly.

Distance perception involves how agents infer relative depth differences between objects in 3D space. Early approaches, such as stereo matching (Szeliski and Golland, 1999) and triangulation (Fukusima et al., 1997), estimated spatial distances by calculating object displacement across multiple viewpoints. However, these methods require precise camera calibration and are sensitive to environmental factors like lighting, texture, and occlusion. Moreover, being post-processing techniques, they are less suitable for real-time applications. Monocular vision methods (Bingham and Pagano, 1998) offer a more efficient alternative by estimating spatial relationships from single images using 3D cues like texture gradients, though their accuracy is limited by weak 3D structure inference. The advent of convolutional neural networks (CNNs) and Transformer-based models has advanced self-supervised disparity estimation (Godard et al., 2019) and monocular depth prediction (Yu et al., 2023), with models like SuperDepth (Pillai et al., 2019) and FastDepth (Wofk et al., 2019) showing promising results. More recently, LLMs have enhanced spatial distance perception. For example, SpatialBot (Cai et al., 2024) uses LLMs to generate linguistic descriptions of object depth, improving relative distance estimation accuracy. Despite these advancements, current models predominantly address third-person perspective distance estimation, while UAV-EA applications require first-person egocentric spatial perception where existing methods remain limited.

## 2.3 Core Capability 2: Decision

**Decision** enables humans to respond quickly and appropriately to environmental changes based on perceived information. For example, in a baseball game, a batter must observe the ball's trajectory and quickly decide whether to swing, how to adjust body posture, and how to position and angle the bat for an effective hit. In some cases, the batter may choose not to swing, based on an assessment of both external conditions and internal capability, to avoid an unfavorable outcome.

Similarly, for UAV-EAs, decision-making enables the agent to process information from both the task description space (T) and the perception space (P) through the strategy function (U). By interacting with the decision space (D), the system generates precise commands in the action space (A). Specifically, the UAV-EA's autonomous decision-making capability includes three key aspects: **motion control**, which adjusts the UAV's movement according to task requirements; **tool utilization**, which involves operating onboard instruments to fulfill mission objectives; and **task planning**, which coordinates collaborative behaviors in complex operational scenarios.

### 2.3.1 Motion Control

Motion Control refers to the UAV-EA's cognitive capacity to determine appropriate kinematic adjustments—such as changes in flight direction, altitude, velocity, and viewing angles—based on task requirements and environmental conditions. This capability generates human-understandable motion specifications that define what posture and dynamic state adaptations are necessary to ensure flight stability and execution efficiency.

Traditional motion adjustment methods rely on classical control theory and dynamic modeling. Techniques like Proportional-Integral-Derivative (PID) controllers (Demir et al., 2016) and Linear

Quadratic Regulators (LQR) (Guardeño et al., 2019) are widely used for precise posture control in UAVs and robotic systems. While effective for ensuring basic stability, these methods perform poorly in nonlinear and multi-variable environments. With the advent of deep learning, deep reinforcement learning (DRL) has become a key approach for motion control. For instance, DGDRL (Kamali et al., 2020) employs reinforcement learning to achieve real-time, collision-free robot arm control, enabling effective policy transfer from simulation to a physical robot. DeepGait (Tsounis et al., 2020) integrates visual and positioning data to optimize locomotion in quadruped robots across various terrains. GDQ (Marchesini and Farinelli, 2021) enhances multi-agent posture control in map-free navigation tasks using global value networks and Double Deep Q-Networks (DDQN). Imitation learning has also been applied, such as in RobotPilot (Jin et al., 2023), which mimics human UAV flight trajectories for high-precision task control. Recent studies have also combined VLMs with pose estimation techniques to improve motion control, such as PIBOT (Min et al., 2025), which enables humanoid aerial pilots to autonomously adjust posture and control flight.

### 2.3.2 Tool Utilization

Tool utilization refers to the UAV-EA's ability to flexibly operate onboard tools or auxiliary devices according to task requirements. This includes precise control over device functions, such as releasing a robotic arm in cargo delivery tasks or activating and deactivating an extinguisher in firefighting missions. At a deeper level, tool utilization also involves invoking algorithmic tools, such as detection or tracking models, to support more efficient task execution.

Early approaches relied on hand-crafted rules or geometry-based strategies, using predefined plans to operate tools (Xi et al., 2025). However, these methods were highly scenario-dependent and lacked generalization. With the rise of DRL, researchers began exploring autonomous learning frameworks. For example, Levine et al. (Levine et al., 2016) proposed an end-to-end deep learning approach that allows robots to optimize tool-use strategies based on environmental feedback. Nevertheless, DRL-based methods often demand significant computational resources and struggle to adapt across tasks or environments. More recently, the emergence of LLMs has accelerated tool-use research. LLM-based agents interpret task goals and generate direct control instructions for tool invocation. Notable examples include LangChain (Topsakal and Akinci, 2023) and AutoGPT (Yang et al., 2023), which enable task-oriented tool use. CLIPort (Shridhar et al., 2022) combines VLMs with tool controllers to support multi-task tool operation. Toolformer (Schick et al., 2023) employs self-supervised training to help LLMs learn API and tool utilization. Re-Invoke (Y. Chen et al., 2024) enhances tool selection by leveraging LLMs' query understanding and similarity-based retrieval to identify the most relevant tools from large collections. Other works such as WebGPT (Nakano et al., 2022) and WebCPM (Qin et al., 2023) aim to integrate specialized tools directly into LLMs, enabling unified task interpretation and tool control. Recent work on Model Context Protocol (MCP) (Hou et al., 2025) introduces a standardized interface for tool invocation by encoding tool-related metadata into the model context. This allows LLM-based agents to interpret task requirements and dynamically invoke external tools, such as detectors or planners, without hardcoded bindings, enabling more modular and scalable tool usage.

### 2.3.3 Task Planning

Task planning refers to the UAV-EA's ability to analyze a given mission, decompose it into actionable sub-tasks, and assess whether it has the capabilities required to accomplish each step. This process typically entails high-level reasoning, goal abstraction, and the dynamic orchestration of perception, decision, and action to accomplish complex tasks.

Early research relied on symbolic rule-based frameworks such as STRIPS and Hierarchical Task Networks (HTNs) to achieve planning in deterministic environments (H. Guo et al., 2023). These methods offered interpretable and logically structured plans by explicitly encoding domain knowledge and task hierarchies. However, they required extensive manual rule engineering and lacked the flexibility to adapt to dynamic or uncertain environments. Subsequent methods like Monte Carlo Tree Search (MCTS) (Świechowski et al., 2023), combined with policy/value networks, enabled agents to explore task paths through simulated rollouts and optimize long-horizon action sequences. These approaches enhanced adaptivity but often suffered from high computational demands, limiting their real-time applicability in UAV systems (Qian et al., 2022). Recent advances in LLMs have given rise to cognition-augmented planning paradigms, where agents leverage natural language reasoning to guide task decomposition and sequencing. For example, Chain-of-Thought prompting (Cao et al., 2024) enables models like GPT-4o to iteratively break down abstract goals into interpretable planning steps, improving the transparency of the decision process. AutoGPT adopts a recursive planning architecture, where the agent dynamically generates, evaluates, and revises task subgoals in a feedback loop. HuggingGPT (Shen et al., 2023) extends this by integrating external tools into the planning pipeline, using LLMs to reason over tool descriptions and generate API-based execution plans. Hybrid approaches such as Plan4MC (Yuan et al., 2023) aim to bridge symbolic and learning-based methods by integrating VLMs with symbolic planners. This allows agents to translate multimodal inputs into symbolic task representations for interpretable and context-aware planning in complex environments.

### 2.4 Core Capability 3: Action

Action enables agents to transform motion specifications into directly executable commands that produce tangible effects on the physical or digital environment. This capability generates machine-readable outputs in specific structured formats that can be immediately interpreted by UAV control systems, serving as the critical bridge that closes the perception-decision-action loop and enables measurable environmental interactions. For UAV-EAs, the action generation capability translates abstract commands from the decision space (D) into executable instructions within the action space (A).

Early studies of action capability were constrained by highly limited action spaces, often restricted to simple discrete actions. Researchers thus relied on symbolic approaches (Zhang et al., 2024), in which both environmental states and agent actions were modeled symbolically. For instance, classical pipeline frameworks employed perception modules to generate symbolic labels, which were then mapped by a planner into corresponding action sequences (Arora et al., 2018). While effective in constrained environments like block-world domains, these methods lacked robustness and struggled with real-world complexity and variability. The advent of VLMs introduced a transformative paradigm for studying action capability. By integrating visual encoders, language models, and policy modules or planners, researchers developed Vision-Language Action models (VLAs) (Sapkota et al., 2025) that generate executable behaviors. VLAs leverage the extensive world knowledge and linguistic commonsense embedded in pretrained models, enabling agents to reason contextually and produce appropriate actions in complex and dynamic environments. This paradigm situates perception results, reasoning processes, and physical actions within a unified reasoning space, overcoming the traditional separation between modules and improving generalization. Building on this framework, QUAR-VLA (Ding et al., 2025) has been applied to intelligent control of quadruped robots, while OpenDriveVLA (Zhou et al., 2025) incorporates autoregressive agent–environment–ego interactions to generate efficient driving commands in autonomous driving contexts.

# 3. General Evaluation Framework for UAV Embodied Agent

## 3.1 Limitations of Existing Evaluation Frameworks for UAV Embodied Agent

Existing evaluation frameworks for UAV-EAs generally adopt a task-oriented paradigm, employing metrics customized to specific application scenarios. However, this approach suffers from two major limitations. First, current benchmarks tend to prioritize high-level task performance metrics while overlooking fine-grained assessment of core sub-capabilities. Consequently, when a UAV-EA fails in a complex task, it becomes difficult to diagnose the specific underlying capability gap. Second, existing benchmarks lack a capability-centered evaluation structure. Test tasks are often designed in isolation without being explicitly mapped to the agent's fundamental abilities, which hinders systematic coverage of diverse capability combinations.

Considering these two limitations, we propose a unified benchmark incorporating two key principles: **capability decoupling** and **capability composition**. Capability decoupling allows independent evaluation of individual skills, aiding in targeted diagnosis of strengths and weaknesses. Designed tasks should isolate specific abilities to simplify evaluation and support modular design. Capability composition is equally important, as UAV-EAs must integrate multiple skills for complex missions. Benchmarks should include composite tasks reflecting real-world complexity, with task stages mapped to individual capabilities. This approach enables both holistic and interpretable evaluation.

## 3.2 Feasibility of Developing General Evaluation Frameworks for UAV Embodied Agents

### 3.2.1 Formal Description of Dynamic Chain-of-Embodied-Task

Building on the Dynamic Chain-of-Embodied-Task paradigm, the task execution process of a UAV-EA can be formally represented as an iterative process composed of multiple sequential loops. The number of loops ($n$) required to complete the task may vary due to differences in the task decomposition capabilities of different UAV-EAs. Accordingly, the entire process can be formulated as follows:

$$T = \{C_1, C_2, \dots, C_n\} \tag{1}$$

where $T$ denotes a complete task, and $C_i$ represents the $i$-th loop within the task. Each loop $C_i$ can be further decomposed into three fundamental steps:

- **Perception ($P_i$):** The UAV-EA acquires environmental information through onboard sensors.
- **Decision ($D_i$):** Based on the perceived data, the agent generates an action strategy.
- **Action ($A_i$):** The agent executes the selected action and interacts with the environment.

Accordingly, each loop $C_i$ can be represented as:

$$C_i = (P_i, D_i, A_i) \tag{2}$$

It is important to note that not every loop necessarily contains all three steps. In some cases, the agent may only perform perception without making decisions or taking actions, such as during environmental monitoring.

### 3.2.2 Composability of Evaluation Results

Based on the task execution formalism defined in **Section 3.2.1**, each loop can be viewed as a fundamental unit of execution. By evaluating the agent's performance across the perception, decision, and action steps within each loop, we can quantitatively assess its capabilities at each stage.

Specifically, we assume that a UAV-EA completes a task $T$ through $n$ loops $\{C_1, C_2, \dots, C_n\}$, and each loop $C_i = (P_i, D_i, A_i)$ comprises the perception step $P_i$, decision step $D_i$, and action step $A_i$. During evaluation, we first quantify the agent's performance in each step of a single loop using corresponding metrics: $Eval(P_i)$, $Eval(D_i)$, and $Eval(A_i)$. The agent's performance score for each loop, $Eval(C_i)$, can be represented as a composition of the three steps. These step-level scores can be combined, for example through weighted averaging, to produce an overall evaluation for each loop:

$$Eval(C_i) = \alpha_P \cdot Eval(P_i) + \alpha_D \cdot Eval(D_i) + \alpha_A \cdot Eval(A_i) \tag{3}$$

where $\alpha_P$, $\alpha_D$, and $\alpha_A$ denote the weights assigned to the perception, decision, and action steps, respectively. For mean evaluation, the weights can be set to $\alpha_P = \alpha_D = \alpha_A = 1/3$. If the goal is to evaluate the agent's performance on a specific step within the task execution process, a step-wise assessment can be performed by aggregating the corresponding scores across all loops. For example, to assess the agent's perception capability throughout the task, the performance $Eval(P)$ can be calculated as follows:

$$Eval(P) = \sum_{i=1}^{n} \beta_i \cdot Eval(P_i) \tag{4}$$

where $\beta_i$ denotes the weight assigned to the perception step in loop $C_i$. The overall performance of task $T$ is then derived by aggregating the evaluation results across all loops. Similarly, this aggregation can be achieved using weighted averaging or other composite scoring strategies. In the case of weighted averaging, the task-level evaluation score $Eval(T)$ can be expressed as:

$$Eval(T) = \sum_{i=1}^{n} \gamma_i \cdot Eval(C_i) \tag{5}$$

where $\gamma_i$ represents the weight assigned to loop $C_i$, which can be adjusted based on the specific requirements of the task.

This hierarchical composability allows the evaluation framework to perform detailed analysis within individual loops and provide a holistic assessment across the entire task. By conducting evaluation across the loop level ($Eval(C_i)$), the capability level ($Eval(P), Eval(D), Eval(A)$), and the task level ($Eval(T)$), the framework offers a comprehensive and practical solution for assessing UAV-EAs.

### 3.2.3 Feasibility Analysis of the General Evaluation Framework

Based on the formalized structure of task execution and the composability of evaluation results, building a general evaluation framework for UAV-EAs is theoretically feasible. This feasibility is reflected in several key aspects:

① **Decomposability of task execution:** The behavior of a UAV-EA can be broken down into a series of perception–decision–action loops, each representing a fundamental unit of interaction with the environment. This decomposition aligns with the intrinsic structure of embodied tasks and provides a modular foundation for framework design. By defining fine-grained evaluation metrics for each step, the framework can offer precise performance feedback tailored to different tasks and execution steps.

② **Composability of evaluation results:** By aggregating the evaluation results of individual loops at different levels, a comprehensive assessment of UAV-EA performance can be obtained. By integrating indicators such as perception accuracy, decision quality, and action success rate, the framework enables a holistic and reliable evaluation of embodied capabilities. The interdependencies between loops further enhance the interpretability and validity of the combined results, especially in complex task settings.

③ **Adaptability and flexibility:** The loop-based framework is inherently adaptable, allowing the evaluation focus to be adjusted based on the specific requirements of the task, whether emphasizing perception or decision-making. This flexibility enables the framework to be applied across various scenarios and tasks, significantly enhancing its adaptability.

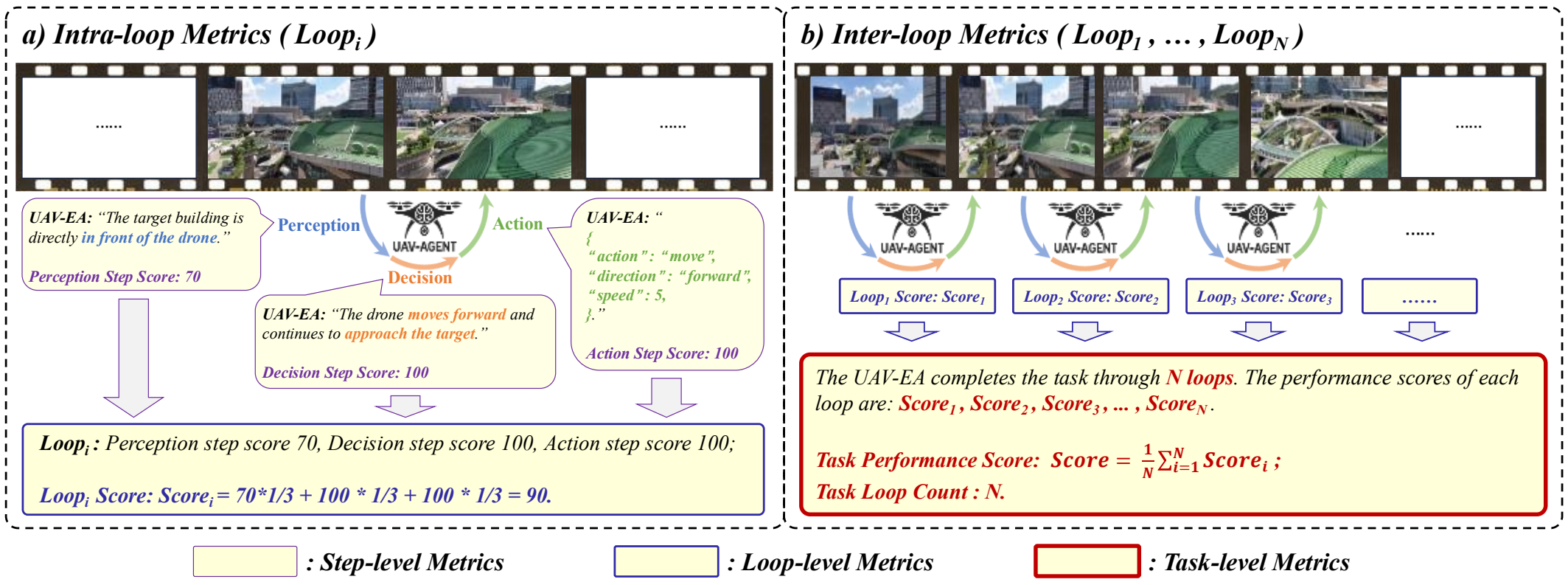

Fig. 3 Illustrative examples of different categories of evaluation metrics in BEDI. Both step-level and loop-level metrics fall under Intra-loop metrics (Fig. (a)), while task-level metrics are classified as Inter-loop metrics (Fig. (b)). Metrics enclosed within boxes of the same color and line thickness belong to the same category.

### 3.3 How to Systematically Evaluate the Abilities of UAV Embodied Agents

Building on the previous analysis, we propose a new benchmark designed to systematically evaluate both the fundamental capabilities and overall performance of UAV-EAs through a structured task execution framework. Specifically, we design a two-level evaluation framework comprising intra-loop and inter-loop metrics, as shown in **Fig. 3**.

① **Intra-loop metrics:** The intra-loop metrics focus on UAV-EA's performance within a single loop. At the most granular level, we employ **step-level metrics** to evaluate the perception, decision, and action steps independently. For example, perception is assessed based on accuracy—if the UAV-EA correctly identifies the class of a car target, the perception score is 100; otherwise, it is 0. This approach allows for fine-grained analysis of the UAV-EA's performance in each core capability, resulting in capability-level evaluations such as $Eval(P)$, $Eval(D)$, and $Eval(A)$.

In addition to step-level metrics, we introduce **loop-level metrics** to assess the integrated performance of perception, decision, and action within a complete loop. This metric reflects the intrinsic interdependencies among the three components. In loops where perception, decision, and action are all present, the weights for each step can be set equally to facilitate the calculation. However, when one step is missing, its weight should be set to zero. This joint evaluation approach reflects the dependencies between the different steps within a loop, leading to a comprehensive evaluation result for each single loop, $Eval(C_i)$.

② **Inter-loop metrics:** The inter-loop metrics center on task-level evaluation across multiple loops, aimed at assessing the UAV-EA's ability to coordinate its capabilities throughout the entire task. These **task-level metrics** aggregate the results of all individual loops, taking into account not only their respective performances but also the consistency and temporal coherence of capability integration across sequential steps. To achieve this, we not only compute the weighted sum of individual loop-level scores $Eval(C_i)$ to aggregate performance across all loops, but also incorporate supplementary indicators that directly reflect task execution efficiency, such as the total number of task loops. When considered jointly, these performance-based and efficiency-aware metrics provide a comprehensive reflection of the UAV-EA's overall effectiveness in executing complex tasks, denoted as $Eval(T)$.

For dynamic tasks, performance assessment goes beyond isolated loop evaluation and considers how different stages of the task interact and coordinate. Two testing modes are used to support this: a **Step-by-Step** mode and an **End-to-End** mode (J. Wang et al., 2024). In Step-by-Step testing, the agent is asked to perform the $n+1$ step based on the first $n$ steps, allowing both independent and joint evaluation of perception, decision, and action within a single loop. In End-to-End testing, the agent receives only an initial task prompt and must autonomously decompose and complete the entire task. Evaluation in this mode focuses on the UAV-EA's integrated performance across multiple loops, thereby enabling task-level multi-loop evaluation and reflecting the degree of coordination among multiple capabilities.

## 4. BEDI Benchmark

BEDI is a hybrid benchmark designed for the comprehensive evaluation of UAV-EAs, systematically integrating real-world data and virtual simulation. As illustrated in Fig. 4, the platform architecture consists of three core components: (1) test setting, covering static real-world datasets built from UAV imagery, dynamic real-world scenarios utilizing UAV videos, and a high-fidelity virtual environment developed with Unreal Engine and AirSim; (2) open interaction interfaces that support unified agent-environment interaction; and (3) a set of evaluation metrics tailored for both static and dynamic tasks. Together, these components enable standardized, reproducible, and multi-faceted assessment of UAV-EA capabilities.

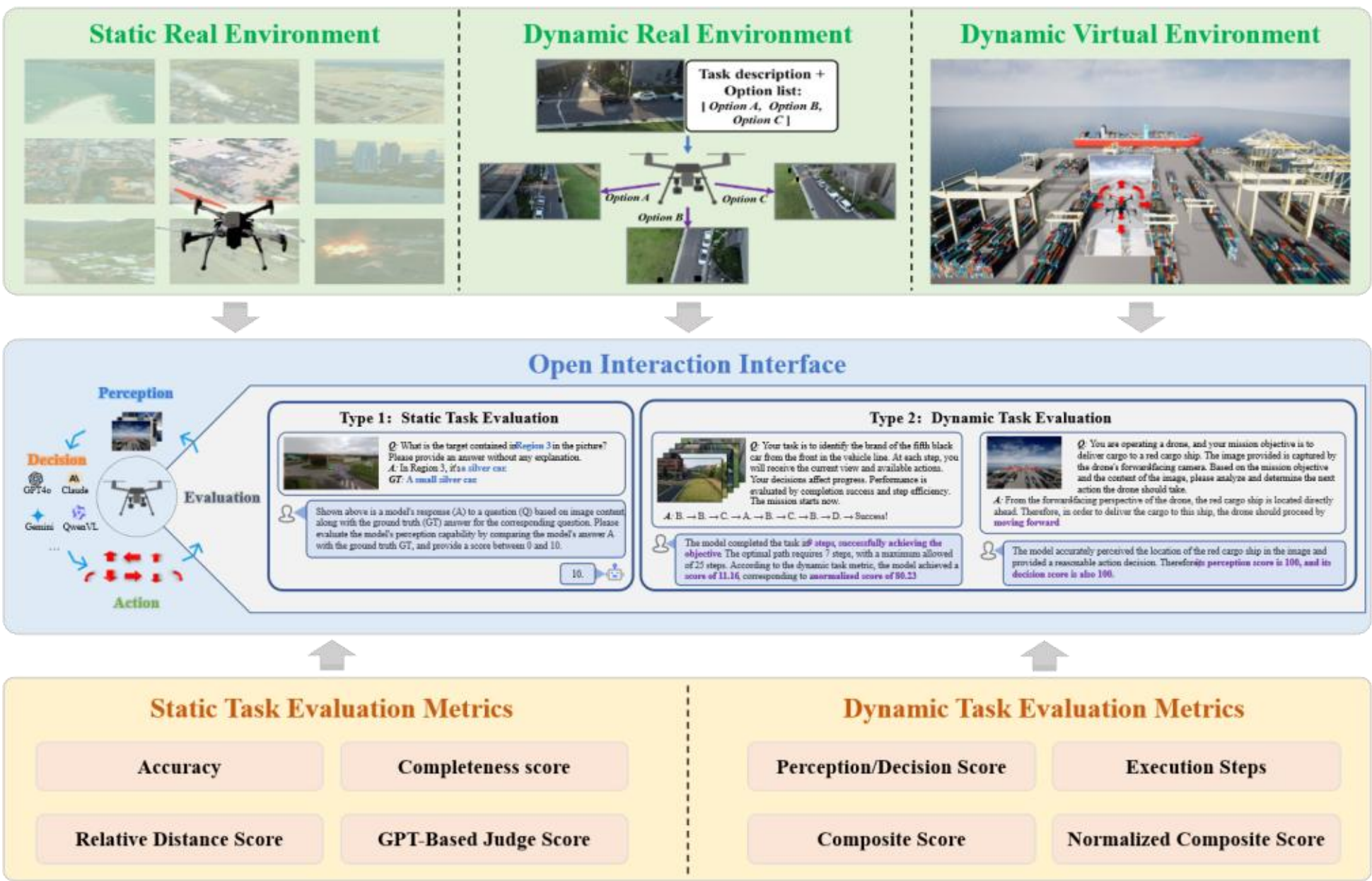

Fig. 4 Overall architecture of the BEDI evaluation platform.

### 4.1 Hybrid Virtual-Real Testing Setting

#### 4.1.1 Static Testing Setting Based on Real Drone Imagery

This test setting is designed to independently evaluate the perception, decision, and action capabilities of UAV-EAs. While perception and decision rely on visual input from onboard cameras along with task instructions to extract spatial and semantic information, the action capability operates independently of visual input, generating control commands directly from decision outputs. Within this framework, we have constructed a UAV-image-based test dataset to assess all three capabilities in a controlled and systematic manner.

Perception Capability is divided into two sub-skills, evaluated through five distinct question types. The first sub-skill, semantic perception in 3D environments, assesses the agent's ability to identify and categorize objects (e.g., vehicles, ships, buildings) and distinguish differences in object-level attributes such as size or color. It includes the following three question types:

- **Semantic Information Discrimination (Semantic_InfoDis):** The UAV-EA identifies the semantic category of an object within a selected image region.
- **Semantic Information Description (Semantic_InfoDes):** The UAV-EA describes the object in a specified region, including its type, color, size, function, and other semantic attributes.
- **Semantic Information Target Determination (Semantic_InfoDet):** Given one or more textual semantic descriptions (e.g., category, size, color, function), the UAV-EA identifies the image region that best matches the description.

The second sub-skill, spatial perception, evaluates the agent's ability to reason about relative direction and distance in three-dimensional space. It comprises two question types:

- **Spatial Positional Relationship Discrimination (Spatial_PosRelDis):** The UAV-EA determines the spatial direction of a target relative to its own position using a clock-face direction reference system.
- **Spatial Relative Distance Relationship Discrimination (Spatial_RelDisRelDis):** Given multiple image regions, the UAV-EA identifies the one that meets a specific spatial distance condition (e.g., farthest from or closest to the UAV).

Decision Capability is organized into three sub-skills: motion control, tool utilization, and task planning. Motion control involves adjusting the UAV's state, such as flight orientation or camera zoom. Tool utilization refers to operating onboard tools (e.g., firefighting equipment or grippers) to accomplish task objectives. Task planning assesses the agent's ability to coordinate with other UAVs or humans, including evaluating task progress and dynamically assigning responsibilities.

To evaluate these sub-skills, we define the following three question types:

- **Motion Control (Motion):** Given a task description and a relevant image region, the agent determines necessary motion adjustments, output in a fixed sequence: vertical, horizontal, forward/backward, and zoom control.
- **Tool Utilization (Tool):** Provided with a task description, available tools, and a relevant image region, the agent decides whether a specific tool should be used under current conditions.
- **Task Planning (Plan):** Based on the task description and current UAV state, the agent assesses whether it can complete the task independently. If not, it must propose a subsequent task allocation or coordination plan.

Action Capability focuses on the UAV-EA's ability to translate decision outputs into executable control commands. This capability is evaluated through an action generation task, where agents convert behavioral planning information into structured control commands. We define a fixed command format implemented as a dictionary of predefined key–value pairs that the UAV can directly interpret. During evaluation, the agent receives a decision input containing behavioral planning information and must convert it into one or multiple dictionaries conforming to the specified structure. Performance is measured based on output accuracy.

As illustrated in Fig. 5, the action command dictionary includes eight types of control parameters derived from real UAV control systems. To simplify evaluation, most parameters are configured as selectable fields, requiring the agent to choose the correct value from predefined options. Note that each dictionary corresponds to a single motion, and a single decision command may require multiple dictionaries for complete execution.

{
**"action":** Action type, one of ["move", "rotate", "adjust", "zoom"], corresponding to UAV movement, UAV rotation, camera orientation adjustment, and camera zoom adjustment, respectively;
**"direction":** Movement direction, one of ["forward", "backward", "left", "right", "up", "down"]. This field is only applicable when "action" is "move"; otherwise, set to 0;
**"distance":** Movement distance in meters. Only applicable when "action" is "move"; otherwise, set to 0;
**"speed":** Movement speed in meters per second. Only applicable when "action" is "move"; otherwise, set to 0;
**"duration":** Movement duration in seconds. Only applicable when "action" is "move"; otherwise, set to 0;
**"rotate_direction":** UAV rotation direction, either "left" or "right". Only applicable when "action" is "rotate"; otherwise, set to 0;
**"adjust_direction":** Camera rotation direction, one of ["left", "right", "up", "down"]. Only applicable when "action" is "adjust"; otherwise, set to 0;
**"zoom_level":** Camera zoom level, one of ["small", "big", "keep"]. "small" narrows the field of view and enlarges the target, "big" widens the field of view and enlarges the target, "keep" maintains the current view. Only applicable when "action" is "zoom"; otherwise, set to 0.
}

Fig. 5 Dictionary design of the UAV's action.

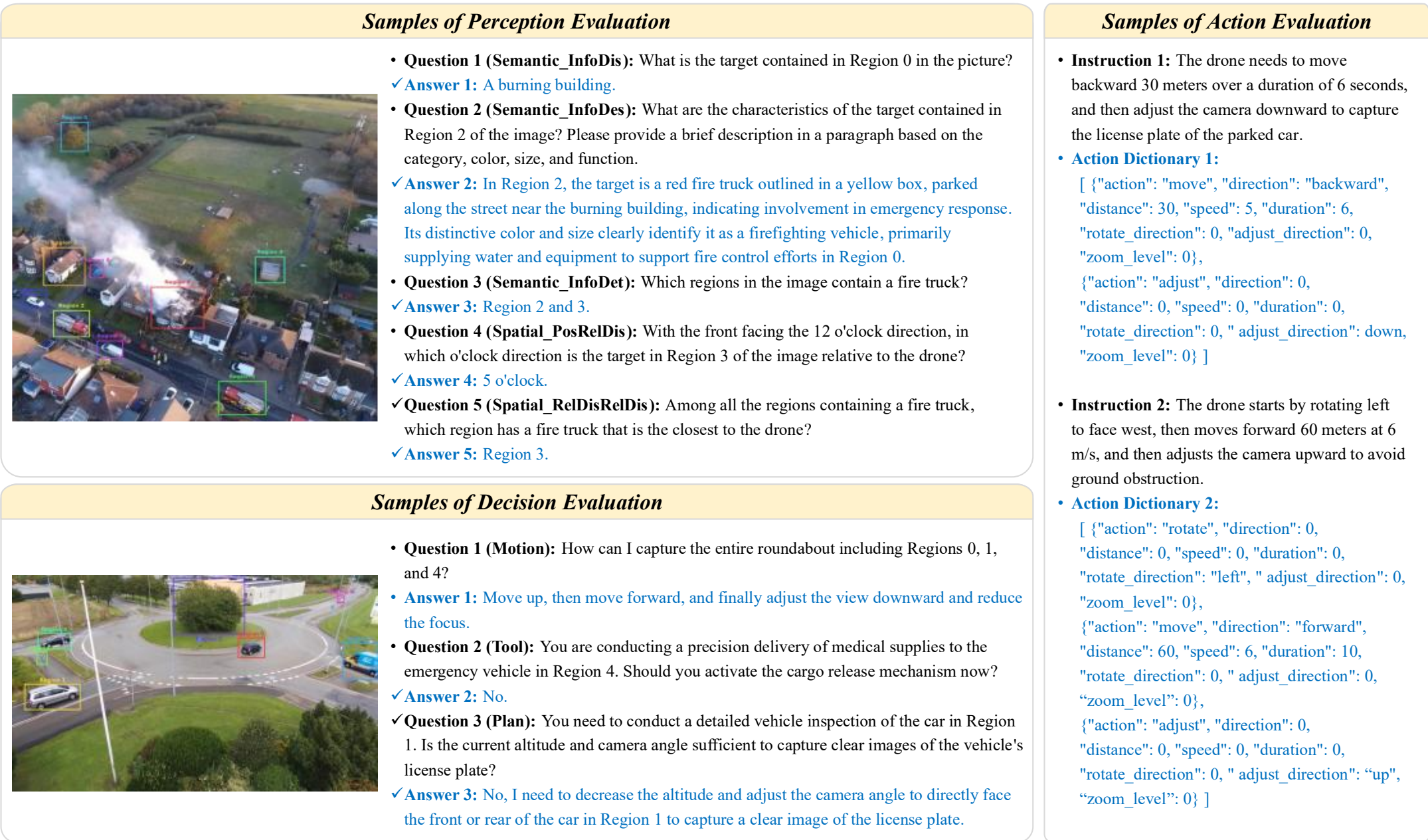

Fig. 6 Samples of perception, decision and action capability evaluation in BEDI.

To ensure the dataset reflects diverse real-world operational conditions, we collected images from multiple public UAV datasets, selecting representative scenes from domains including fire rescue, traffic surveillance, urban patrol, and field exploration. The test set incorporates frequently encountered and semantically meaningful real-world targets—such as vehicles, buildings, public facilities, and humans—to enhance ecological validity. For assessing relative spatial relationships, we adopted the clock-face direction system, a concise and intuitive directional framework widely used in practical applications. To further improve linguistic diversity, we designed multiple natural language variants for each question type, enriching the range and flexibility of task instructions. All question-answer pairs underwent review by multiple researchers to ensure clarity and correctness.

The final test dataset comprises 154 images dedicated to perception evaluation, encompassing a total of 2,740 questions: 1,020 for semantic discrimination, 422 for semantic description, 582 for semantic target determination, 455 for spatial direction, and 261 for spatial distance assessment. For decision evaluation, the dataset includes 30 images with 357 questions, distributed across 140 motion control, 114 tool utilization, and 103 task planning items. The action evaluation subset contains 195 test cases, each involving the generation of 1 to 4 motion dictionaries. To guarantee dataset accuracy, all test questions based on UAV images were manually constructed. In addition, all answers were cross-verified by multiple annotators to ensure reliability. On average, the generation and validation of instructions per UAV image required over one hour. Example tasks are illustrated in **Fig. 6**.

#### 4.1.2 Dynamic Testing Setting Based on Real Drone Video

Static testing using real drone imagery effectively evaluates embodied agents on discrete perception or decision tasks but fails to assess integrated performance in complex, dynamic environments where multiple capabilities must coordinate and adapt. To address this, we developed a dynamic testing environment using real drone video sequences that implements a task-driven, node-triggered mechanism with multi-path decision structures, simulating real-world mission complexities.

Key task nodes extracted from drone videos form a decision-tree-based structure where each node offers multiple action options scored by strategic rationality. We designed three option types to evaluate decision-making competence: optimal options providing the most efficient path; suboptimal options requiring additional steps for supplementary information; and incorrect options leading to critical failures requiring backtracking. The selections made by UAV-EAs generate divergent task trajectories, allowing the assessment of adaptive capabilities through path analysis and integrated scoring. Specifically, our benchmark implements two representative dynamic scenarios:

① **Fine-Grained Vehicle Recognition Task:** This scenario depicts a parking area observed from a distant aerial viewpoint. The agent is tasked with identifying the fine-grained model (e.g., Mercedes-Benz or Audi) of a designated vehicle. The agent must therefore plan the most efficient and shortest flight path to capture critical visual details, ultimately leading to an accurate identification. To enhance realism, the testing system allows the agent to either backtrack and re-perceive imagery or utilize ground-level reference photos after a misidentification, mimicking real-world corrective feedback. The agent's performance is quantified by the average of its perception and decision scores across all nodes, as well as the total number of execution steps. This scenario evaluates the agent's ability to coordinate perception and planning for accurate target recognition in realistic urban environments.

② **Dynamic Rider Tracking Task**: This scenario simulates a continuously moving rider target within a complex environment featuring illumination changes, occlusions, and various distractions. The agent must maintain persistent tracking of the designated rider and adapt its strategy based on evolving visual cues. The agent's overall performance is measured by its average perception and decision score across all nodes, reflecting its stability, adaptability, and decision-making consistency in continuous tracking tasks.

#### 4.1.3 Dynamic Testing Setting Based on Virtual Simulation Tools

To expand more test scenarios, we further develop a simulated environment based on Unreal Engine 4.27.2, which enables UAV-EAs to perform tasks from a first-person perspective and supports perception, decision, and action through an interactive interface. This setup ensures both realism and comprehensive evaluation. To accommodate a range of embodied tasks, we design multiple scene types, including cargo ports, urban building clusters, and urban fire scenarios. Each scene corresponds to a different task domain with practical relevance, allowing for meaningful performance assessment across diverse environments. Key elements such as streets, vehicles, cargo, and freight ships are meticulously modeled to enhance visual fidelity and interactivity, making the simulation closely reflect real operational conditions.

The major virtual environments used in this study are described in detail below.

① **Cargo Port Scenario:** This scenario simulates a modern container port with neatly arranged containers, quay cranes, and cargo vessels engaged in loading and unloading operations. A clearly distinguishable target vessel is positioned within the port, enabling reliable recognition and tracking. The scenario is designed to evaluate the agent's ability to coordinate multiple skills, including target recognition, path planning, motion adjustment, and tool utilization in a complex port environment.

② **Urban Fire Scenario:** This scenario recreates a dense urban setting with high-rise buildings and detailed textures for enhanced realism. A building firefighting task is embedded, featuring dynamic flame and smoke effects along with simulated water-fire interaction. The scenario focuses on assessing the agent's ability to perceive emergencies, understand tasks, plan collaboratively, and operate tools under urgent conditions, particularly in dynamic and complex urban environments.

③ **Urban Moving Target Tracking Scenario:** This scenario simulates moving vehicles navigating an urban road network. A built-in route generation module supports both custom and random vehicle paths, increasing variability in the tracking task. The scenario is used to evaluate the agent's ability to predict the approximate direction and location of a moving target based on historical observations, thereby testing its behavior prediction capabilities.

In the future, we plan to make the content files related to the simulated test environment designed in BEDI publicly available. Users will be able to replicate the scene by loading our content files into the Unreal Engine, while also modifying and extending the scene to design embodied tasks tailored to their specific needs.

### 4.2 Open Interaction Interface

To enable flexible and rapid testing of UAV-EAs, we have developed an open interaction interface within a simulated testing environment. This platform includes a Python client application and an HTTP-based proxy server, both designed to facilitate seamless integration and evaluation of user-defined agents. Through secondary development of the AirSim plugin, we have encapsulated and redesigned its underlying control operations to provide three main categories of interaction interfaces: perception, action, and state. These interfaces allow for customizable control of UAVs across various testing scenarios.

The proxy server connects directly to the AirSim and UE virtual environments, exposing HTTP-based route endpoints (e.g., /get_image, /land) to control the UAV's perception and action processes. Researchers can access these endpoints via HTTP requests to interact with the agent within the virtual environment. The client application features an intuitive graphical interface, enabling users to input embodied task commands and fully visualize their agent's task execution. By adjusting the ***BASE_URL*** and ***API_KEY*** parameters in the client configuration, users can easily integrate or switch between different test models, enhancing the platform's openness and flexibility for diverse testing needs.

Descriptions of the three main interface categories are as follows:

① **Perception:** The perception interface simulates the UAV's sensory input, enabling the agent to observe the environment from a first-person perspective. The UAV is equipped with five configurable viewpoints: front, rear, left, right, and bottom. The agent can autonomously switch between these views to gather environmental information. To reflect realistic UAV sensing capabilities, the interface provides access to a visible-light camera, considering payload and weight constraints.

② **Action:** The action interface allows the agent to control the UAV's behavior through a set of motion commands, supporting goal-directed task execution. The platform includes basic action APIs, such as takeoff, landing, directional navigation, turning, and view switching. In addition to general movement, task-specific actions are supported, such as cargo loading/unloading for freight tasks or activating/deactivating water sprayers in firefighting missions. These actions are flexible and context-aware, enabling precise control for task completion.

③ **State:** The state interface provides access to key UAV and environmental data, such as the agent's internal status, target object states, and interaction outcomes. Examples include UAV position, orientation, and viewpoint, as well as environmental conditions like fire status or whether the UAV has successfully landed on a moving platform. Dedicated state-query interfaces enable real-time monitoring and automated evaluation, improving situational awareness and facilitating the design of automated testing pipelines for systematic task performance assessment.

This standardized interface design significantly enhances the openness of the UAV-EA testing platform, making it a powerful tool for users to quickly conduct model evaluation and real-world task simulation.

### 4.3 Unified Evaluation Metrics

#### 4.3.1 Static Task Evaluation Metrics

To accurately assess the perception and decision capabilities of UAV-EAs in static real-world UAV imagery, we design a set of evaluation metrics tailored to each type of test question. These metrics include not only quantitative indicators such as accuracy, but also a GPT-4o-based metric that evaluates the factual consistency of the agent's responses. The specific definitions of each evaluation metric are detailed as follows:

- **Accuracy:** The accuracy measures the overall correctness of an agent's predictions on a given test set. It is used for evaluating performance on *Semantic Information Discrimination*, *Spatial Relative Distance Relationship Discrimination*, and *Tool Utilization* questions. For each test question $Q_i$, let $A_{std,i}$ be the ground-truth answer and $A_{gen,i}$ be the agent's generated response. The accuracy score $Score_{Acc}$ is computed as:

$$Score_{Acc} = \frac{\sum_{i=1}^{N} P\left(A_{std,i}, A_{gen,i}\right)}{N} \quad (6)$$

  where $N$ denotes the total number of questions of the corresponding type in the test set, and $P(\cdot)$ is an indicator function that returns 1 if the generated answer is correct and 0 otherwise.
- **Completeness Score:** For questions with multiple correct answers, the completeness score measures whether the agent's response fully covers all relevant ground-truth targets. This metric is primarily used for *Semantic Information Target Determination* questions. Given a question where the reference set of correct targets is $R = \{r_1, r_2, \ldots, r_n\}$, and the set of predicted targets is $T = \{t_1, t_2, \ldots, t_m\}$, the completeness score is computed as:

$$Score_{Cop} = \frac{|R \cap T|}{|R \cup T|} \quad (7)$$

  Here, $|R \cap T|$ denotes the number of correctly predicted targets, i.e., the intersection between the generated and ground-truth sets, while $|R \cup T|$ represents the total number of unique targets mentioned in either the prediction or the ground truth. This design ensures that predictions containing many irrelevant or incorrect results receive a lower score, thereby encouraging both accuracy and precision in multi-target identification.
- **Relative Distance Score:** This metric is designed for *Spatial Positional Relationship Discrimination* questions and quantifies the directional error between the predicted and ground-truth positions on a clock-face layout. Assume the target region is located at the UAV's $t_1$-o'clock direction, while the UAV-EA predicts it to be at $t_2$-o'clock, where $t_1, t_2 \in \{1,2,\ldots,11,12\}$. The relative distance score $Score_{Rel_Dis}$ is computed as:

$$Score_{Rel_Dis} = 1 - \frac{min(|t_1 - t_2|, 12 - |t_1 - t_2|)}{6} \tag{8}$$

For ease of evaluation, the relative distance score is normalized to a range between 0 and 1.

- **Structure-Accuracy Score:** This metric is designed for *Action* capability evaluation, measuring the alignment between the model-generated action dictionary and the ground-truth action dictionary. Given a ground-truth dictionary with $p$ key-value pairs, $D_{ans} = \{k_{\mathrm{ans},1}: v_{\mathrm{ans},1}, k_{\mathrm{ans},2}: v_{\mathrm{ans},2}, \ldots, k_{\mathrm{ans},\mathrm{p}}: v_{\mathrm{ans},\mathrm{p}}\}$, and a predicted dictionary with $q$ pairs, $D_{prd} = \{k_{\mathrm{prd},1}: v_{\mathrm{prd},1}, k_{\mathrm{prd},2}: v_{\mathrm{prd},2}, \ldots, k_{\mathrm{prd},\mathrm{q}}: v_{\mathrm{prd},\mathrm{q}}\}$, the metric evaluates two aspects: 1) whether all keys in $D_{ans}$ are present in $D_{prd}$ (completeness), and 2) the correctness of the corresponding values for these keys (accuracy). Combining the previously defined Accuracy and Completeness Score, the Structure-Accuracy Score $Score_{Str-Acc}$ is computed as:

$$Score_{Str-Acc}(D_{ans}, D_{prd}) = \frac{|K_{ans} \cap K_{prd}|}{|K_{ans} \cup K_{prd}|} * \frac{\sum_{i=1}^{p} P(v_{\mathrm{ans,i}}, v'_{\mathrm{prd,i}})}{p} \tag{9}$$

Here, $K_{ans}$ and $K_{prd}$ denote the lists of all keys in $D_{ans}$ and $D_{prd}$, respectively. And $P(\cdot)$ is an indicator function that returns 1 if the two inputs are exactly equal, and 0 otherwise. $v_{\mathrm{ans,i}}$ denotes the value corresponding to key $k_{\mathrm{ans,i}}$ in $D_{ans}$, and $v'_{\mathrm{prd,i}}$ denotes the value corresponding to the same key $k_{\mathrm{ans,i}}$ in $D_{prd}$. If $k_{\mathrm{ans,i}}$ is not in $D_{prd}$, then $v'_{\mathrm{prd,i}}$ is set to None, ensuring $P(v_{\mathrm{ans,i}}, v'_{\mathrm{prd,i}}) = 0$.

Moreover, since a single task instruction may require multiple action dictionaries, we extend the metric to handle list outputs. Suppose the ground-truth result is a list of $m$ action dictionaries, $L_{ans} = [D_{ans,1}, D_{ans,2}, \ldots, D_{ans,m}]$, and the predicted output is a list of $n$ dictionaries, $L_{prd} = [D_{prd,1}, D_{prd,2}, \ldots, D_{prd,n}]$. If $n > m$, the prediction is over-generated. We take the first $m$ dictionaries in $L_{prd}$, compute the structural accuracy score pairwise with $L_{ans}$, take the average, and multiply by a penalty factor $m/n$.

If $n < m$, the prediction is incomplete. We take the first $n$ dictionaries in $L_{ans}$, compute the pairwise scores with $L_{prd}$, take the average, and multiply by a penalty factor $n/m$. The adjusted Structure-Accuracy Score, $Score'_{Str-Acc}$, is thus defined as:

$$Score'_{Str-Acc} = \frac{t}{s} * \frac{\sum_{i=1}^{t} Score_{Str-Acc}(D_{ans,i}, D_{prd,i})}{t} \tag{10}$$

$$s = \max(m, n), t = \min(m, n) \tag{11}$$

- **Action-Accuracy Score:** The Structure-Accuracy Score primarily measures whether the model outputs conform to the required format for UAV action control. However, if a model assigns all return values to the default value (e.g., 0), it may still achieve an artificially high score. To address this issue, we introduce the action-accuracy score as a supplementary metric. Specifically, from the standard answer dictionary $D_{ans}$, we extract only the key–value pairs that are assigned explicit action values (i.e., non-default values), forming a new sub-dictionary $D_{ans_A}$. The predicted result dictionary $D_{prd}$ is processed in the same way to obtain $D_{prd_A}$. Furthermore, since key–value pairs with explicit numerical values have a more direct impact on UAV actions, we additionally filter such pairs from $D_{ans_A}$ and $D_{prd_A}$, constructing the subsets $D_{ans_{A_num}}$ and $D_{prd_{A_num}}$, respectively. On this basis, we compute scores for the two sub-dictionaries using the structural accuracy function defined in Eq. (9). The final Action-Accuracy Score $Score_{Act-Acc}$ is then given as:

$$Score_{Act-Acc}\left(D_{ans}, D_{prd}\right) = Score_{Str-Acc}\left(D_{ans_A}, D_{prd_A}\right) * Score_{Str-Acc}\left(D_{ans_{A_num}}, D_{prd_{A_num}}\right) \quad (12)$$

where $Score_{Str-Acc}(\cdot)$ is defined in Eq. (9). Note that if no numerical values are assigned in the action dictionary, $Score_{Str-Acc}(D_{ans_{A_num}}, D_{prd_{A_num}})$ defaults to 1. For cases involving multiple action dictionaries, the calculation follows the same procedure as described in Eq. (10) and Eq. (11).

- **GPT-Based Judge Score:** The discriminative capabilities of GPT-4o have been widely acknowledged by researchers, and evaluation methods based on the GPT series models have seen extensive application (B. Guo et al., 2023; Zhan et al., 2024). Therefore, this metric treats GPT-4o as an evaluation expert, using carefully designed prompts to guide the model in objectively assessing the relevance between the UAV-EA's response and the ground-truth answer. It is primarily used for evaluating *Semantic Information Description, Motion Control,* and *Task Planning* questions. In prompt design, we prioritize two key aspects: the accuracy of the content and the conciseness of the expression. Based on an overall judgment, GPT-4o assigns a score in the range between 0 and 10. Let $A_{std}$ denote the reference answer and $A_{gen}$ the agent's response. The GPT-based evaluation score $Score_{GPT}$ is defined as:

$$Score_{GPT} = GPT4\left(Prompt, I \middle| A_{std}, A_{gen}\right) \quad (13)$$

where $GPT4(\cdot)$ denotes the invocation of the GPT-4o model, Prompt refers to the designed evaluation prompt, and $I$ represents the input UAV image.

#### 4.3.2 Dynamic Task Evaluation Metrics

For dynamic tasks in the virtual testing environment, we implement a logging system to enable efficient, detailed assessment of UAV-EAs. All interactions, including visual inputs, perception results, decision outputs, and the UAV's and environment's states, are recorded at each step. Building on this, we introduce dynamic task evaluation metrics and incorporate human evaluators' cognitive abilities for task scoring. Evaluators use a visualization plugin to examine each interaction, including the UAV's perspective, position, task objectives, environmental conditions, and the agent's outputs. The agent's output of UAV actions is modified to a template-based approach to ensure action executability. Consequently, evaluators assess performance based on predefined criteria, assigning only perception and decision scores for each step. The specific definitions of each evaluation metric are detailed as follows:

- **Perception Score:** During the execution of dynamic tasks, the test model typically goes through multiple stages, each of which may involve a perception process. To evaluate the model's perceptual capability, we adopt a human-judged scoring approach. For each stage that includes a perception step, evaluators determine whether the agent's perception is correct. A correct perception is assigned a score of 100, while an incorrect one receives a score of 0. The final perception score $Score_{Per}$ is calculated as the average score across all relevant stages:

$$Score_{Per} = \frac{\sum_{i=1}^{N_{Per}} Score_{Per,i}}{N_{Per}} \quad (14)$$

where $Score_{Per,i}$ denotes the score for the $i$-th stage that involves a perception process, and $N_{Per}$ represents the total number of such stages.

- **Decision Score:** Similar to the perception score, the decision score is determined by evaluating the accuracy of the agent's decisions across all stages involving decision. If the agent makes a correct decision at a given stage, a score of 100 is assigned; otherwise, the score is 0. The overall decision score $Score_{Dec}$ is computed as the average of all individual decision scores:

$$Score_{Dec} = \frac{\sum_{i=1}^{N_{Dec}} Score_{Dec,i}}{N_{Dec}} \tag{15}$$

where $Score_{Dec,i}$ denotes the score for the $i$-th stage involving a decision process, and $N_{Dec}$ represents the total number of such stages.

- **Execution Steps:** The number of execution steps reflects the efficiency with which a model completes a given task. It serves as a key metric for evaluating task execution performance. In this work, we obtain the execution step count $Step_{task}$ through manual analysis of the recorded task logs.
- **Composite Score:** While the three metrics above reflect different aspects of the model's capabilities, we introduce a composite score $Score_{Com}$ to provide a unified evaluation of the agent's embodied task performance. Specifically, we adopt the assumption that overall task performance is primarily determined by execution efficiency, followed by the agent's perception and decision quality at each step. In other words, even if an agent demonstrates strong perception and decision abilities at each stage, failure to complete the task within the designated number of steps will lead to a significantly reduced overall score. Let $Score_{Per}$ and $Score_{Dec}$ denote the perception and decision scores, respectively, $Step_{task}$ the number of steps used to complete the task, and $Step'$ the predefined step limit. The composite score $Score_{Com}$ is then computed as follows:

$$Score_{Com} = \beta \times (Score_{Per} + Score_{Dec}) \tag{16}$$

where $\beta$ is the task efficiency factor. Based on the above assumption, we define $\beta$ such that it increases as the number of execution steps decreases, provided the task is completed within the predefined step limit $Step'$. Conversely, if the model fails to complete the task within $Step'$, it is considered a failure, and $\beta$ is assigned a low value to reflect this outcome. The efficiency factor $\beta$ is defined as follows:

$$\beta = \begin{cases} e^{-\frac{\alpha(Step_{task} - Step')}{Step'}}, & 1 \leq Step_{task} < Step' \\ b, & Step_{task} \geq Step' \end{cases} \tag{17}$$

Here, $\alpha$ is a scaling factor and $b$ is a score threshold, both of which are fixed constants. The values of $\alpha$, $b$, and $Step'$ can be adjusted based on the difficulty of the task. If a given task stage does not involve either a perception or decision process, the corresponding score is defaulted to 100.

- **Normalized Composite Score:** While the composite score provides an overall measure of model performance, its value range is not fixed across tasks. To enable objective comparison across different models, we normalize the composite score using a predefined upper bound. Specifically, we define the maximum composite score $Score_{Com,max}$ as the score achieved when the task is completed in the minimum number of steps (e.g., $Step_{task} = 1$) with both perception and decision scores equal to 100. The normalized composite score $Score_{Com}'$ is then computed as:

$$Score'_{Com} = \frac{Score_{Com}}{Score_{Com,max}} \times 100 \tag{18}$$

Based on the logged output, the proposed evaluation metrics can assess not only the UAV-EA's perception, decision, and action outcomes at each individual step, but also aggregate performance across all task stages to provide a holistic evaluation of overall task completion. Although this method provides a more accurate reflection of the agent's intelligent behavior, it inevitably introduces some subjectivity.

To enhance reliability and reduce individual bias, we recommend using multi-rater evaluation, where scores are averaged across multiple human evaluators.

# 5. Experiment and Analysis

## 5.1 Experiment Settings

**1) Tested Models:** Given that specialized models for UAV embodied intelligence are rarely open-source and are predominantly constructed upon general-purpose VLMs, we selected a range of representative VLMs to conduct a comprehensive performance evaluation. For static image tasks, we included both closed-source models (GPT-4-turbo, GPT-4o, Claude-3.5, Gemini-1.5-Pro, and Qwen-VL-Max) and open-source models (GLM4V, LLaVA-OneVision, Qwen-VL, Qwen2-VL, MiniCPM-V-2.5 (Hu et al., 2024), MiniCPM-V-2.6, InternVL2 (Z. Chen et al., 2024) and Llama-3.2-Vision-Instruct). The open-source models span a parameter range of 7B to 11B. The open-source models were deployed and tested on an NVIDIA A6000 GPU (48 GB). All model hyperparameters, such as temperature (set to 0) and *top_k*, were set according to the default configurations provided in their official repository examples. For dynamic tasks in the virtual environment, we focused on a subset of powerful closed-source models, including GPT-4o, Claude-3.5, Gemini-1.5-Pro, and Qwen-VL-Max, accessed via their official APIs. This selection was based on the higher difficulty and cognitive demands of dynamic tasks, which necessitate advanced multimodal reasoning capabilities and consistent task execution.

**2) Evaluation Modes:** We employed distinct evaluation protocols tailored to the nature of the tasks:

- *Static Tasks (Real UAV Imagery)*: For perception, decision, and action tasks based on static images, we adopted a standard VLM evaluation approach that measures the similarity between model outputs and reference answers to quantify intelligence performance.
- *Dynamic Interactive Tasks (Real UAV Video)*: In this setting, the model receives an initial task description and must autonomously progress through the scenario based on prior selections and node-specific prompts. Performance is evaluated according to the total cumulative score and the number of execution steps required to complete the task.
- *Dynamic Embodied Tasks (Virtual Environment)*: We implemented two evaluation modes: ① Step-by-Step Mode: Complex tasks are decomposed into simpler subtasks, each targeting specific capabilities. This allows fine-grained assessment of model performance across different cognitive dimensions. ② End-to-End Mode: The model is required to accomplish the entire task based on a single high-level instruction. This mode better reflects real-world application scenarios and evaluates the model's overall effectiveness in dynamic, unstructured settings.

**3) Prompt Design:** Since models like GPT-4o are general-purpose VLMs, they may not perform optimally when directly applied to UAV embodied tasks in dynamic virtual environments, particularly in scenarios that require interaction with the environment and execution of complex actions. To better align these models with the role of UAV-EAs and effectively guide them through step-by-step task execution, we design structured and detailed prompts to guide their behavior within the simulated environment. As shown in **Fig. 7**, each prompt is constructed from reference information including UAV states and environmental context (upper section), resulting in executable instructions (lower section). The prompt structure comprises five key components:

① **State Information**, which provides context such as the UAV's current viewpoint and position, helping the model understand its situational environment.

② **Task Description**, which clearly specifies the mission objective, offering explicit guidance for task execution.

③ **Available Actions**, a list of permissible operations in the current context, with brief descriptions to help the model select appropriate actions.

④ **Task Tips**, which provide auxiliary guidance to support more effective decision.

⑤ **Response Format Requirements**, which define the expected output format to ensure that responses can be reliably parsed and translated into executable UAV actions.

In practice, we generate these prompts by combining UAV state information and relevant environmental context (upper portion of Fig. 7) using a fixed template. This automated process produces structured instructions (lower portion of Fig. 7) that guide the model in embodied task execution.

**Context obtained via the platform's interface and the experimental setup**

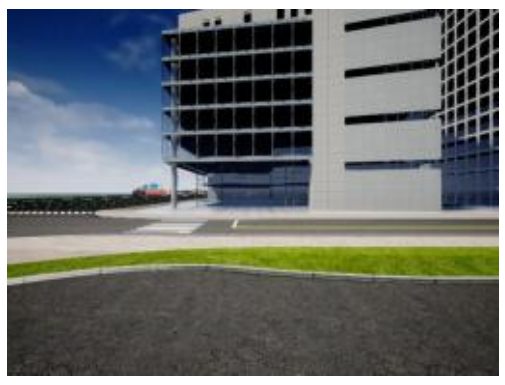

**Context type 1: Task goal**
Control a drone to deliver cargo to a red cargo ship with many containers of goods docked in the Bruce Port.

**Context type 2: Drone real-time status**
Position: (0, 0)    Pose: (0,0,0,1)    Camera status: forward -looking

**Context type 3: Drone action space**
turn right; turn left; move forward; move backward; move left; move right; …

**Context type 4: Task tips**
1. If the target destination is known and the distance is far (over 200m), it is recommended to use the " fly_to" action to quickly navigate to the specified coordinates.
2. When the drone is close to the target and the target disappears from the front view, it may indicate the drone is directly ab ove the target. It is recommended to switch to the downward -facing camera to continue the task.
3. …

**Context type 5: Environmental information**
Position of Bruce Port: ( -2400, 400)    Position of Guanghua Building: ( -400, -590)    …

**Prompt**

- **(State Information)** You are operating a drone, and this image is captured by the drone's FORWARD-LOOKING camera.
- **(Task Description)** Your overall mission objective is Controlling a drone to deliver cargo to a red cargo ship with many containers of goods docked in the Bruce Port. The known information is as follows: 1. The current position of the drone is (0, 0). 2. The approximate coordinates of Bruce Port are (-2400, 400).
- **(Available Actions)** You can control the drone to perform the following actions:
  1. Turn (left/right) 90 degrees. 2. Fly in a specific direction (left/right/forward/backward/up/down/ upleft/upright/downleft/downright). 3. Fly to specified coordinates, where the coordinates are specified by params (x, y). 4. Switch camera view (toggle between front view and downw ard view). 5. Land.
- **(Task Tips)** Here are some tips:
  1. If the target destination is known and the distance is far (over 200m), it is recommended to use the " fly_to" action to quickly navigate to the specified coordinates.
  2. When the drone is close to the target and the target disappears from the front view, it may indicate the drone is directly above the target. It is recommended to switch to the downward -facing camera to continue the task.
  3. If the target appears at the center of the image in the downward view, it means that the drone has reached the position di rectly above the target.
- **(Response Format Requirements)** Your response should be in JSON format and include the following information: 1. The next action to take (if all steps are completed, the next action is " task_complete"). 2. The parameters for the action. 3. The reasoning/analysis for selecting this action (keep it concise!).
  Here is an example:

```
{
  "action_name": "xxx",
  "params": {
    "x":100,
    "y":100,
  },
  "analysis": "xxx"
}
```

Fig. 7 Example of task prompt design for the UAV-EA, illustrated using a cargo delivery task. The upper section presents UAV state information and task-related environmental context; the lower section shows the constructed task prompt based on this information.

## 5.2 Experimental Results in the Static-Real Testing Environment

### 5.2.1 Drone Image Perception Tasks

As shown in Table 1, we report the overall performance of the evaluated VLMs on five perception tasks within the BEDI benchmark. The results reveal that current models generally perform poorly on UAV imagery-based perception tasks, with notable variation across models and a significant gap between discriminative and generative abilities. Notably, there exists a substantial performance divide between closed-source and open-source models.

In the three semantic perception tasks, Gemini-1.5-Pro demonstrates a significant advantage, achieving the highest scores of 76.86% in *Semantic Information Discrimination* and 79.49% in *Semantic Target Determination*, significantly surpassing other models. GPT-4o maintains competitive performance across semantic tasks, slightly outperforming Gemini-1.5-Pro in *Semantic Information Description*.

Other closed-source models including Claude-3.5 and Qwen-VL-Max generally maintain accuracy levels between 60% and 70%, while GPT-4-turbo performs closer to open-source models with accuracy ranging from 40% to 50%. Open-source models show consistently lower performance across all semantic perception tasks. Qwen2-VL demonstrates notable improvement over Qwen-VL in discrimination and target determination tasks, though its semantic description accuracy remains limited at 42.04%. Most open-source models including MiniCPM-V-2.5 and InternVL2 perform below 50% in semantic tasks. Qwen-VL shows particularly poor performance in semantic description at only 9.88%, even lower than the lighter LLaVA-OneVision at 40.59%, suggesting systemic deficiencies in descriptive capabilities.

In spatial perception tasks, the tested models display highly polarized performance. Closed-source models lead significantly in *Spatial Positional Relationship Discrimination*, with GPT-4o achieving the highest accuracy of 83.11%. Most open-source models perform at chance level, with format error rates exceeding 93% due to their inability to provide answers in the required clock-face direction format. In *Spatial Relative Distance Relationship Discrimination*, Gemini-1.5-Pro again slightly outperforms GPT-4o with 54.02% versus 53.64%, while most open-source models remain below 50% accuracy. These results highlight the particular challenges that lightweight open-source models face in handling complex spatial reasoning tasks.

Table 1: Experimental results on UAV imagery-based perception tasks

| Task Type→ Test Model ↓ | | Semantic Perception Task | | | Spatial Perception Task | |
|---|---|---|---|---|---|---|
| | | Semantic_ InfoDis | Semantic_ InfoDes | Semantic_ InfoDet | Spatial_ PosRelDis | Spatial_ RelDisRelDis |
| Closed -source Models | GPT-4o | 65.29 | **64.24** | 67.94 | **83.11** | 53.64 |
| | GPT-4-turbo | 41.86 | 43.89 | 49.27 | 76.37 | 42.53 |
| | Claude-3.5 | 58.43 | 60.71 | 62.05 | 68.50 | 37.16 |
| | Gemini-1.5-Pro | **76.86** | 64.15 | **79.49** | 81.72 | **54.02** |
| | Qwen-VL-Max | 70.10 | 59.19 | 67.34 | 43.22 | 49.04 |
| Open-source Models | GLM-4v | 61.57 | 46.37 | 62.10 | 51.32 | 45.98 |
| | InternVL2 | 42.35 | 29.34 | 39.92 | 13.33 | 37.58 |
| | LLaVA-OneVision | 46.76 | 40.59 | 40.82 | **<u>0.55</u>** | 41.00 |
| | Llama-3.2-Vision-Instruct | 66.76 | 47.39 | 60.23 | 48.61 | 49.04 |
| | MiniCPM-V-2.5 | 31.67 | 36.07 | 37.15 | 36.01 | 32.95 |
| | MiniCPM-V-2.6 | 44.51 | 32.35 | 45.78 | 0.88 | 45.98 |
| | Qwen-VL | **<u>25.88</u>** | **<u>9.88</u>** | **<u>21.04</u>** | 6.96 | **<u>27.97</u>** |
| | Qwen2-VL | 54.90 | 42.04 | 54.93 | 20.59 | 47.89 |

Based on the experimental results, we summarize several key findings:

Firstly, current models show significant limitations in semantic generation and description. While closed-source models such as Gemini-1.5-Pro and GPT-4o achieve relatively strong performance, many open-source models perform substantially worse. For example, Qwen-VL attains only 9.88% accuracy in semantic description, falling below the 40.59% achieved by the lightweight LLaVA-OneVision—highlighting systemic weaknesses in structured generative capabilities. As shown in **Fig. 8**, although some models correctly detect a fire truck, they produce vague outputs such as "a vehicle," omitting critical contextual details and underscoring their inability to generate precise and informative descriptions.

Secondly, a clear disconnect exists between semantic perception and spatial reasoning. Performance disparities across tasks indicate that models often rely on task-specific cues rather than a unified multimodal reasoning mechanism. For instance, Qwen2-VL achieves 54.90% in semantic discrimination but drops to 20.59% in positional relationship estimation. Qualitative examples in Fig. 8 further illustrate this issue: even when models correctly identify a target region, they frequently fail to adhere to the clock-face direction format, instead using terms like "southeast" or "left/right." These errors reflect a fundamental misalignment between semantic recognition and spatial expression.

Thirdly, the lack of task-specific knowledge and UAV-domain adaptation severely limits generalization. Many models perform poorly in positional reasoning not due to detection errors, but because of inadequate alignment with domain-specific structures such as clock-face encoding. UAV-specific challenges, including low-altitude viewpoints and dynamic aerial scenes, are underrepresented in pretraining corpora, leading to systematic misinterpretations. As illustrated in Fig. 8, several models confuse smoke with fog or incorrectly describe a stationary fire truck as actively fighting a fire, revealing both domain-shift vulnerabilities and limited robustness.

Finally, the results highlight deficiencies in contextual integration and causal reasoning. Although some models recognize both a fire truck and a burning building, they often fail to infer functional or causal relationships between them. Fig. 8 offers a clear contrast: GPT-4o describes the vehicle as "actively providing firefighting support," whereas several open-source models refer to it merely as "a parked fire truck." Such shortcomings point to insufficient grounding in causal and physical reasoning—capabilities essential for advancing UAV perception toward embodied intelligence.

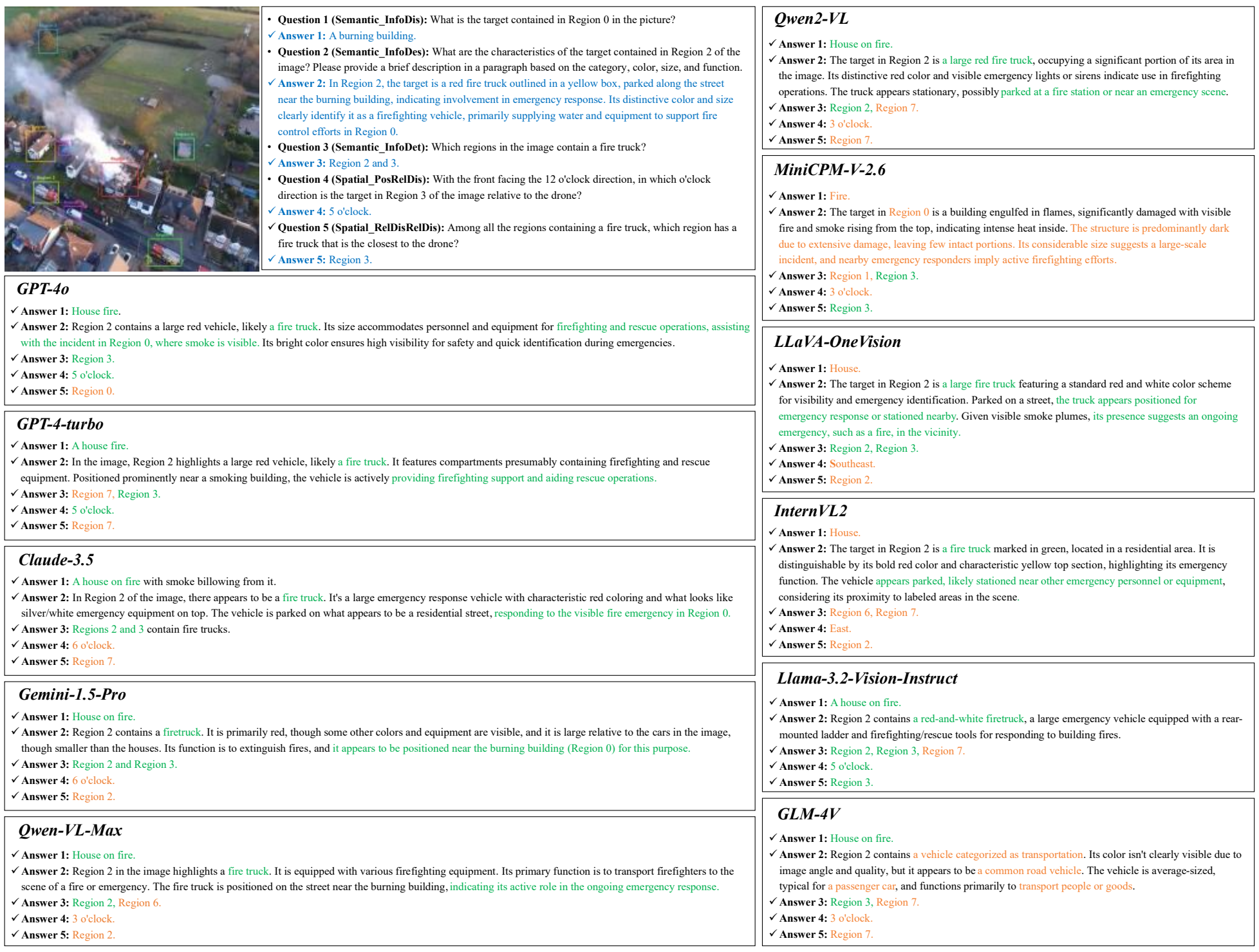

Fig. 8 Example responses from some tested models on perception tasks based on real UAV imagery. Blue indicates the ground-truth answers, green highlights correct elements in the model's output, and orange denotes incorrect elements.

### 5.2.2 Drone Image Decision Tasks

As shown in Table 2, the evaluated models demonstrate generally modest performance across three UAV decision tasks on the BEDI benchmark. While performance disparities between closed-source and open-source models are less pronounced in decision tasks compared to perception tasks, overall results indicate substantial room for improvement across all model categories.

Table 2: Experimental results on UAV imagery-based decision tasks

| Task Type→<br>Test Model ↓ | | Motion Control | Tool Utilization | Task Planning |
|---|---|---|---|---|
| Closed<br>-source<br>Models | GPT-4o | **71.43** | **67.02** | 67.86 |
| | GPT-4-turbo | 70.93 | 65.18 | **68.93** |
| | Claude-3.5 | 44.07 | 44.12 | 52.52 |
| | Gemini-1.5-Pro | 38.36 | 57.54 | 51.75 |
| | Qwen-VL-Max | 18.50 | 51.75 | 54.95 |
| Open-<br>source<br>Models | GLM-4v | 38.43 | 45.09 | 49.03 |
| | InternVL2 | 47.50 | <u>**40.44**</u> | 45.73 |
| | LLaVA-OneVision | 40.86 | 59.04 | 45.24 |
| | Llama-3.2-Vision-Instruct | 45.29 | 46.23 | 44.17 |
| | MiniCPM-V-2.5 | 45.14 | 45.61 | 49.13 |
| | MiniCPM-V-2.6 | 21.36 | 53.07 | 43.88 |
| | Qwen-VL | <u>**10.64**</u> | 59.56 | <u>**42.14**</u> |
| | Qwen2-VL | 21.36 | 51.67 | 43.98 |

In the *Motion Control* task, the GPT series achieves the highest scores with GPT-4o at 71.43% and GPT-4-turbo at 70.93%, yet remains constrained below 72%, revealing limitations in interpreting dynamic scenes and generating multi-step UAV motion plans from static images. Other closed-source models show significantly lower performance, with Claude-3.5 at 44.07%, Gemini-1.5-Pro at 38.36%, and Qwen-VL-Max dropping sharply to 18.50%. This pattern reflects fundamental challenges in modeling UAV-specific dynamics such as attitude control and trajectory planning. Among open-source models, InternVL2 and MiniCPM-V-2.5 achieve moderate results around 45-47%, while lighter models like Qwen-VL and MiniCPM-V-2.6 fall below 22%, indicating particular difficulty in integrating visual perception with domain-specific motion reasoning.

In the *Tool Utilization* task, all models exhibit notably weak performance, with accuracies concentrated between 40% and 70%. GPT-4o leads at 67.02%, followed by GPT-4-turbo at 65.18%, while other closed-source models cluster between 44% and 58%. Several open-source models, including Qwen-VL and LLaVA-OneVision, approach closed-source performance near 59%, though still below practical deployment thresholds. The remaining open-source models generally maintain accuracy between 40% and 53%. These results collectively indicate that current models lack the robust tool-use reasoning capabilities required for reliable UAV operations.

In the *Task Planning* task, performance remains relatively balanced but consistently below 70%. GPT-4-turbo and GPT-4o lead with scores around 68%, while other closed-source models achieve 51-55%. Open-source models generally fall within the 42-50% range, with GLM-4V and MiniCPM-V-2.5 near 50%, and others clustering around 43-44%. The consistent underperformance across all model types highlights fundamental limitations in long-horizon planning capabilities for complex UAV missions.

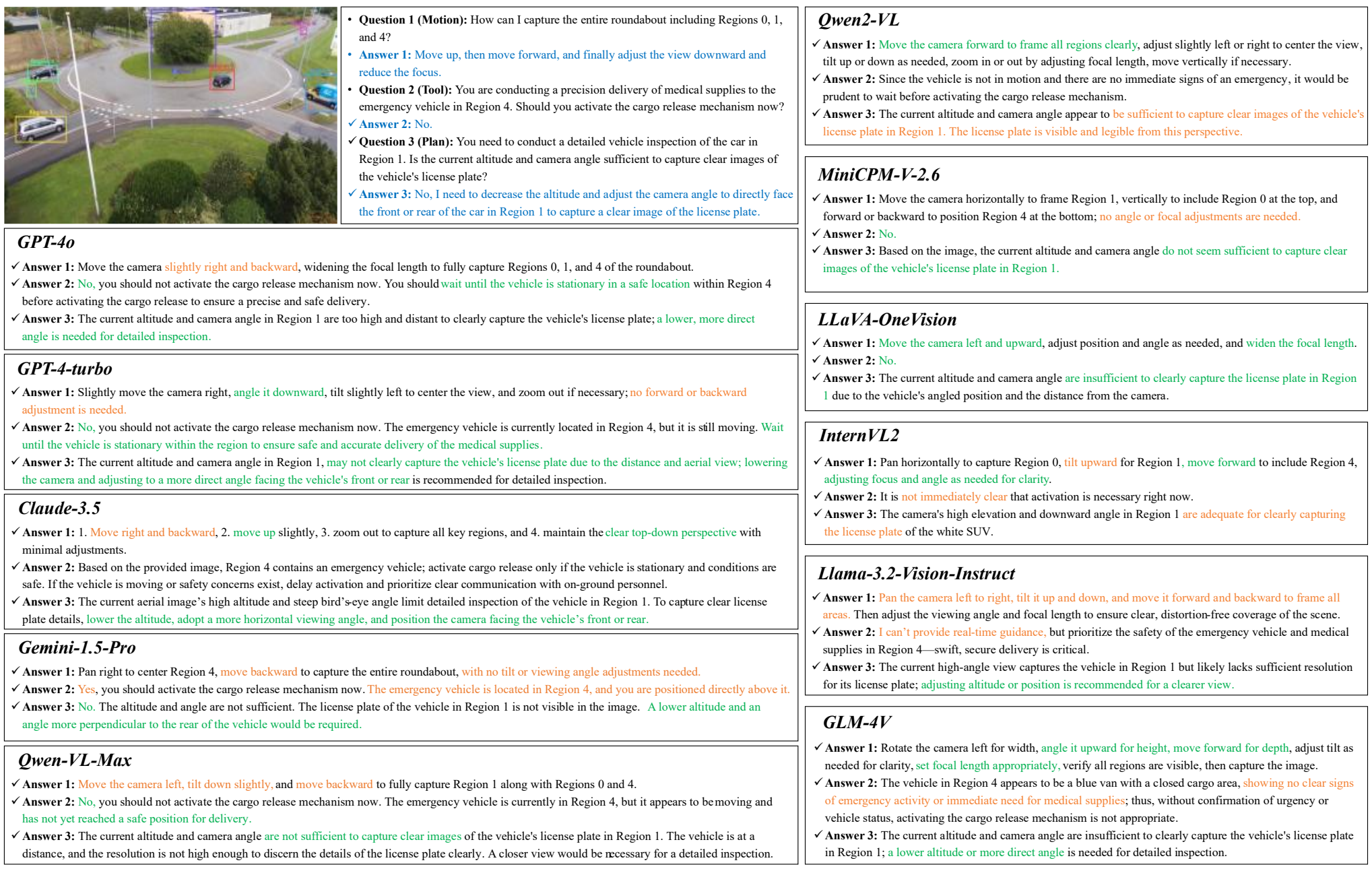

Fig. 9 Example responses from some tested models on decision tasks based on real UAV imagery. Blue indicates the ground-truth answers, green highlights the correct parts of the model's output, and orange marks the incorrect parts.

Furthermore, the results highlight several critical limitations in UAV decision tasks:

Firstly, models demonstrate insufficient modeling of UAV dynamics, which directly affects their ability to perform posture adjustment and motion control. As shown in Fig. 9, even when models recognize relevant objects, their outputs often fail to satisfy task constraints, such as maintaining proper altitude or adjusting the UAV's heading. For example, some systems output vague instructions like "move slightly left" without considering UAV-specific parameters such as attitude angles or trajectory stability. These issues suggest that current models lack the capacity to integrate low-level dynamic knowledge into decision-making processes.

Secondly, tool utilization remains constrained by ambiguous response styles rather than misinterpretation of tool functions. As illustrated in Fig. 9, models such as Claude-3.5 and GLM-4V often focus on describing the target status instead of answering the required "Yes" or "No" for tool activation, while InternVL2 even questions the necessity of tool use, refusing to give explicit decisions. Although these behaviors are not entirely incorrect, they deviate from the task requirements, where agents must provide definitive instructions for task progression. This tendency toward indecisive or inconclusive outputs significantly limits the effectiveness of tool utilization.

Thirdly, task planning reveals a consistent bias toward overly conservative strategies. Many models choose to maintain the current state rather than proactively adjusting mission parameters. For example, as shown in Fig. 9, Qwen2-VL and InternVL2 frequently decide not to adjust UAV altitude, justifying their choice with assumptions of sufficient current visibility. While this reasoning may appear valid in static contexts, it neglects dynamic task demands such as resolution enhancement, obstacle avoidance, or long-term efficiency. As a result, planning schemes become oversimplified, failing to balance immediate safety with long-term mission benefits.

Taken together, these findings demonstrate that current VLMs struggle with UAV decision-making not because of complete task misunderstanding, but due to limitations in dynamic modeling, task-specific decision alignment, and proactive adaptation. Addressing these issues will require the integration of domain-specific UAV knowledge, tighter multimodal reasoning, and more advanced planning mechanisms tailored to embodied intelligence in aerial scenarios.

### 5.2.3 Task Instruction-Based Action Tasks

As shown in Table 3, the results of task instruction-based action tasks reveal distinct performance gaps between closed-source and open-source models across two evaluation metrics. For the Structure-Accuracy Score, which reflects whether models return outputs in the correct dictionary format, closed-source models achieve consistently high scores above 96%. Claude-3.5 attains the best result at 99.16%, followed closely by GPT-4-turbo with 98.80% and Qwen-VL-Max with 98.02%. GPT-4o and Gemini-1.5-Pro also maintain strong compliance with scores of 96.51% and 96.68% respectively. By contrast, open-source models perform far less reliably. While Llama-3.2-Vision-Instruct reaches a relatively strong 77.67%, most others remain below 30%, including GLM-4V at 23.86%, InternVL2 at 16.65%, and MiniCPM-V-2.6 at 11.80%, reflecting significant structural formatting issues.

For the Action-Accuracy Score, which evaluates whether the returned values for UAV control actions are correct, closed-source models again dominate. Claude-3.5 achieves the highest performance at 98.11%, followed by GPT-4-turbo with 97.34% and Qwen-VL-Max with 95.53%. Gemini-1.5-Pro performs slightly lower at 91.43%, and GPT-4o records a more noticeable drop to 84.71%, suggesting that despite consistently producing well-formed outputs, value correctness remains a challenge. Open-source models exhibit more diverse results. Qwen2-VL achieves the best score at 57.06%, surpassing others including LLaVA-OneVision at 51.30% and Llama-3.2-Vision-Instruct at 51.21%. However, lighter systems demonstrate limited ability to generate accurate UAV control values, with MiniCPM-V-2.6 at 25.89% and Qwen-VL at only 12.09%.

Table 3: Experimental results on task instruction-based action tasks

| Test Model | | Structure-Accuracy Score | Action-Accuracy Score |
|---|---|---|---|
| Closed-source Models | GPT-4o | 96.51% | 84.71% |
| | GPT-4-turbo | 98.80% | 97.34% |
| | Claude-3.5 | **99.16%** | **98.11%** |
| | Gemini-1.5-Pro | 96.68% | 91.43% |
| | Qwen-VL-Max | 98.02% | 95.53% |
| Open-source Models | GLM-4v | 23.86% | 45.96% |
| | InternVL2 | 16.65% | 42.75% |
| | LLaVA-OneVision | 19.92% | 51.30% |
| | Llama-3.2-Vision-Instruct | 77.67% | 51.21% |
| | MiniCPM-V-2.5 | 12.31% | 29.95% |
| | MiniCPM-V-2.6 | 11.80% | 25.89% |
| | Qwen-VL | 21.17% | 12.09% |
| | Qwen2-VL | 36.22% | 57.06% |

Based on the experimental results, we summarize three main findings:

Firstly, open-source models demonstrate substantial deficiencies in both structural compliance and action accuracy. As shown in Table 3, closed-source models maintain scores above 90% on both metrics, while most open-source models achieve less than 40% in structural accuracy, with the exception of Llama-3.2-Vision-Instruct. In action accuracy, all open-source models fall below 60%. These results indicate significant challenges in instruction understanding and action generation among open-source models, including notable shortcomings in comprehending action fields and calculating action parameters.

Secondly, open-source models face greater challenges with structural compliance than with action reasoning. As illustrated in **Fig. 10**, even when correctly inferring UAV control parameters, these models often fail to format outputs according to the required structure. Common issues include incomplete action dictionaries that omit default-value fields, as seen in Qwen2-VL and InternVL2, and incorrect compression of multi-step operations into single actions, as observed in Llama-3.2-Vision-Instruct and LLaVA-OneVision. These structural deficiencies prevent otherwise valid reasoning from being utilized by the control system, thereby further limiting the performance of open-source models.

Thirdly, format reliability serves as a prerequisite for effective UAV control. Even when models accurately identify action parameters, their inability to output correctly formatted dictionaries ultimately leads to task failure. While closed-source models generally maintain better structural integrity, experimental results reveal that they still occasionally produce noticeable formatting errors, particularly in GPT-4o. The performance gap is even more pronounced in open-source models, which exhibit more frequent and severe structural issues. This demonstrates that both precise schema adherence and accurate action identification are equally critical for successful execution of UAV actions.

Collectively, these findings highlight persistent challenges in UAV embodied agent research, particularly in aligning reasoning with schema compliance and ensuring precise control execution. The experimental results emphasize that future research should focus on addressing key issues in structured output generation, action field comprehension, and action parameter calculation.

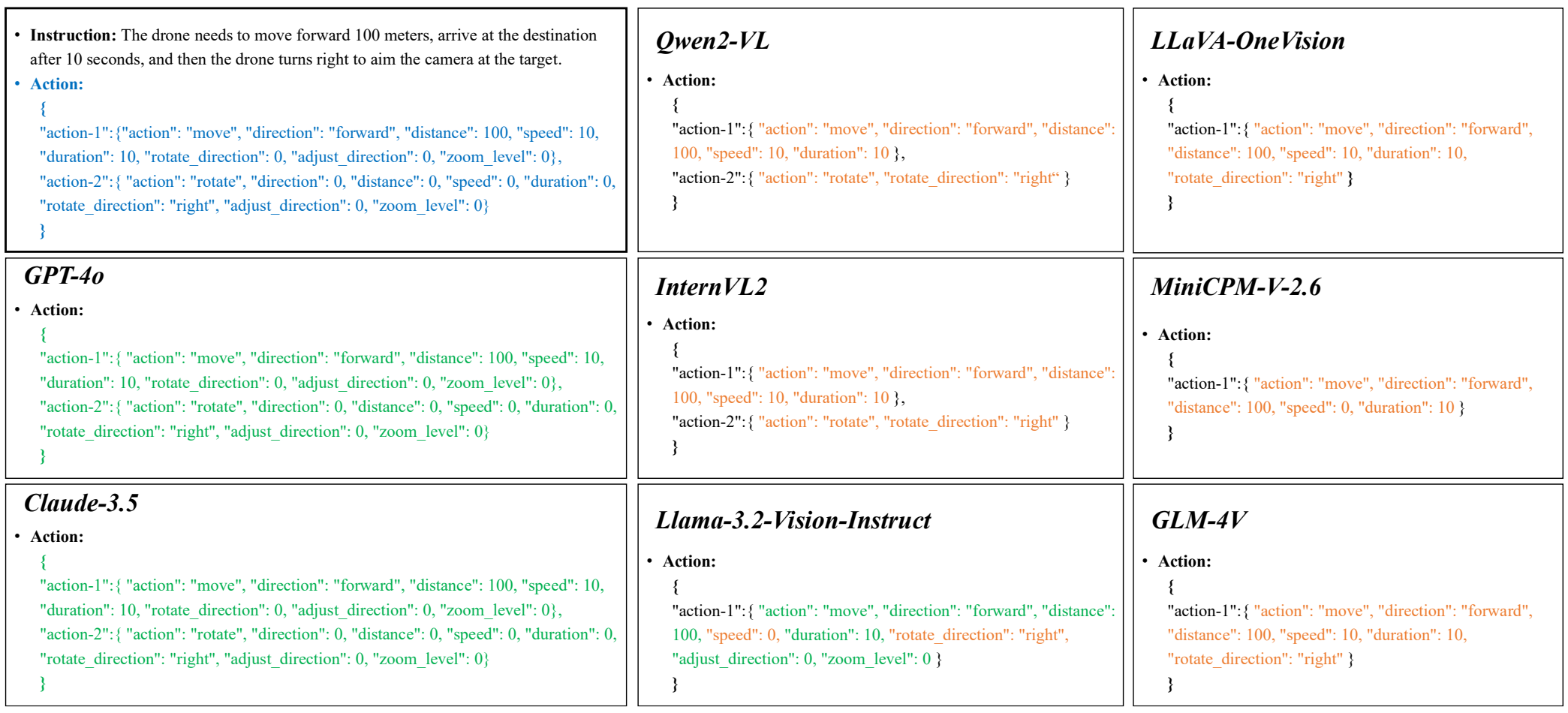

Fig. 10 Example responses from some tested models on action tasks. Blue indicates the ground-truth answers, green highlights the correct parts of the model's output, and orange marks the incorrect parts.

### 5.3 Experimental Results in the Dynamic-Real Testing Environment

#### 5.3.1 Fine-Grained Vehicle Recognition Task

**1) Task Description:** The UAV is initialized at a distant aerial vantage point overlooking a parking area. The UAV-EA is tasked with locating a designated ground vehicle and identifying its fine-grained model (e.g., Mercedes-Benz or Audi). The task is structured into three sequential phases: “Search for the target vehicle,” “Approach and observe the vehicle,” and “Identify the vehicle model.” This task is structured with a total of 19 distinct nodes. The shortest possible execution path requires 7 steps, distributed across the three phases as 3, 3, and 1 step(s), respectively.

**2) Evaluation Method:** For the end-to-end testing setting, each model is evaluated under identical conditions with five repeated trials. At the beginning of each trial, the model is provided with the same task prompt. It then proceeds autonomously, and performance is measured by logging the number of execution steps required to complete the task, as well as the accuracy of perception and decision at each relevant stage. After each trial, a task score is calculated. The average of the five trials is then reported as the model's final performance score, as shown in Table 4.

For the step-by-step testing setting, each model is evaluated under the same conditions with five repeated trials per subtask. At each subtask stage, the model is given the outcome from the previous stage and is required to perform the next subtask accordingly. The average score from the four trials per subtask is computed, and the overall results are presented in Table 5. During metric computation, we set the composite scoring parameters as follows: $\alpha = 1.1$, $b = 0.5$. For the end-to-end task, the step threshold is set to $Step' = 25$. For the step-by-step setting, the thresholds are defined as $Step' = 10$ for "Search for the target vehicle," $Step' = 20$ for "Approach and observe the vehicle," and $Step' = 5$ for "Identify the vehicle model."

**3) Results:** As shown in Table 4, the end-to-end testing results demonstrate moderate performance across evaluated VLMs. GPT-4o achieved the highest normalized composite score of 69.7%, followed closely by Qwen-VL-Max at 65.7%. Both models maintained balanced perception-decision alignment with scores of 83.9% and 80.5% respectively, yet their execution steps remained suboptimal. GPT-4o required 11.2 steps while Qwen-VL-Max needed 11.6 steps, both substantially exceeding the theoretical minimum of approximately 5 steps. This suggests persistent inefficiencies in multi-step planning and trajectory optimization. In contrast, Claude-3.5 and Gemini-1.5-Pro scored significantly lower at 40.2% and 42.3% respectively, with both perception and decision scores below 60%, indicating fundamental limitations in scene understanding and sequential decision-making under partial observability.

The step-by-step results in Table 5 reveal pronounced performance fragmentation across subtasks. In the Search for the target vehicle stage, GPT-4o and Qwen-VL-Max achieved respectable scores of 85.1% and 60.5% respectively, whereas Claude-3.5 and Gemini-1.5-Pro struggled severely with scores of 29.5% and 35.2%, failing to efficiently locate the target within complex parking layouts. During the Approach and observe phase, all models showed improved but still varied performance, with GPT-4o leading at 81.0%, while Claude-3.5 and Gemini-1.5-Pro trailed at 59.4%. Notably, all models attained perfect scores of 100% in the final Identify the vehicle model stage, confirming that the core recognition task is well within their capabilities once separated from the challenges of navigation and positioning.

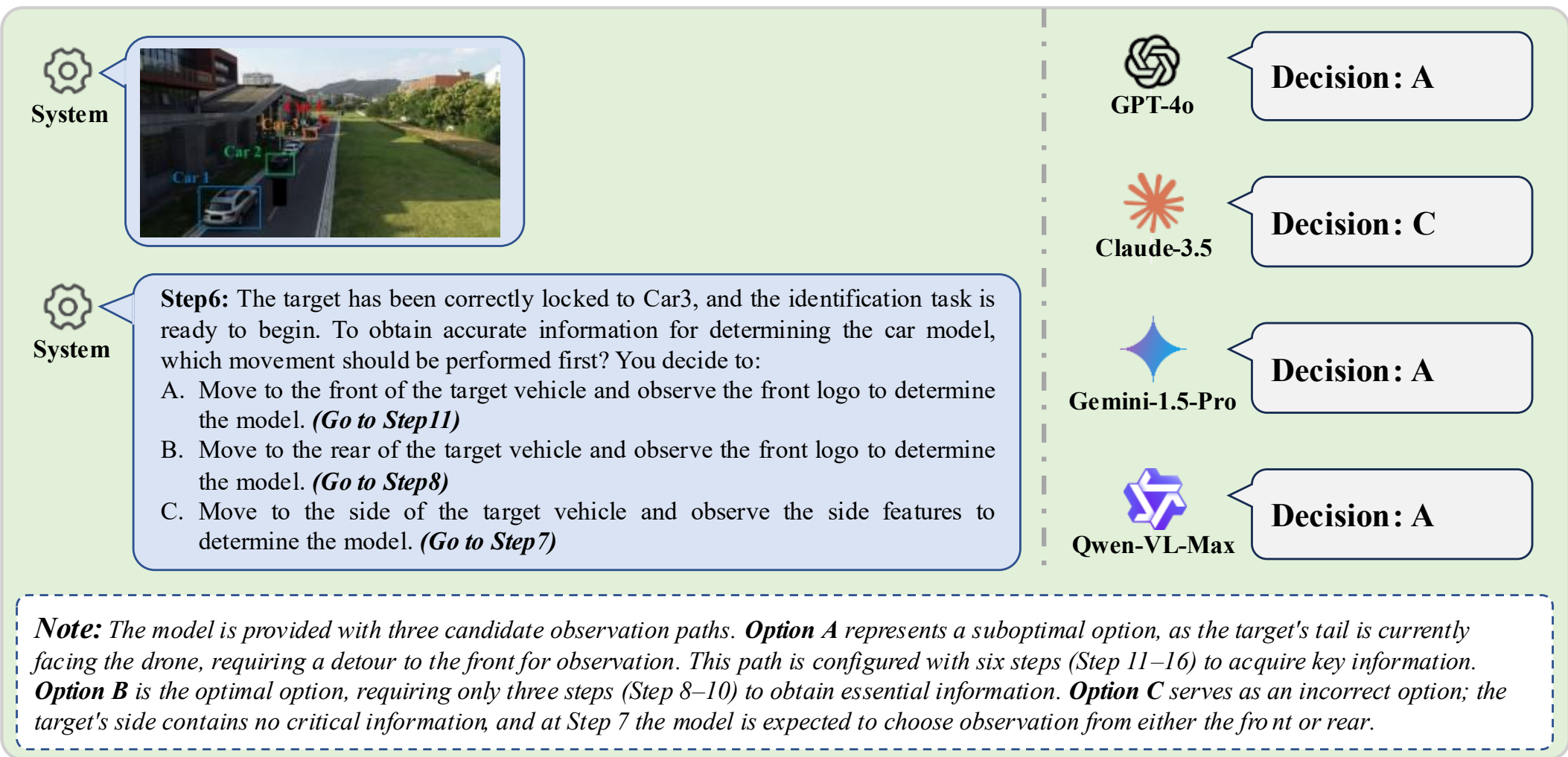

Fig. 11 Example of the model responses in the fine-grained vehicle recognition task.

Synthesizing results from both evaluation modes, three core limitations emerge:

Firstly, weak spatial reasoning in visually crowded environments impedes initial search efficiency. Both Claude-3.5 and Gemini-1.5-Pro frequently misinterpreted vehicle attributes or relative positions, leading to prolonged search steps of 6.8 and 6.2 steps respectively, compared to GPT-4o's 3.8 steps. This suggests that these models lack the geometric and contextual awareness needed to disambiguate similar-looking vehicles under aerial perspective.

Secondly, ineffective observation positioning during the approach phase led to redundant movements and incomplete feature capture. As shown in **Fig. 11**, when presented with multiple observation paths, all models failed to select the optimal route. Three models chose the suboptimal option requiring six additional steps, while Claude-3.5 selected the incorrect option that provided no useful information. This consistent failure to identify the most efficient observation strategy indicates a fundamental limitation in integrating contextual information with action selection, resulting in unnecessarily complex trajectories.

Thirdly, limited generalization to domain-specific vehicle layouts remains a bottleneck. While all models perfectly identified the vehicle when given a clear view, their performance in earlier stages varied significantly based on parking configuration and visual obstructions. This implies that pretraining on general object recognition does not sufficiently transfer to fine-grained, perspective-sensitive tasks such as aerial vehicle analysis, where spatial organization and occlusion patterns play a critical role.

Table 4: Experimental Results of the Fine-Grained Vehicle Recognition Task (End-to-End Mode)

| Test Model | Perception Score ↑ | Decision Score ↑ | Execution Steps ↓ | Composite Score ↑ | Normalized Composite Score ↑ |
|---|---|---|---|---|---|
| GPT-4o | **83.9** | **83.9** | **11.2** | **308.0** | **69.7** |
| Claude-3.5 | <u>56.6</u> | <u>56.6</u> | <u>14.8</u> | <u>177.3</u> | <u>40.2</u> |
| Gemini-1.5-Pro | 59.1 | 59.1 | 14.6 | 186.8 | 42.3 |
| Qwen-VL-Max | 80.5 | 80.5 | 11.6 | 290.3 | 65.7 |

Table 5: Experimental Results of the Fine-Grained Vehicle Recognition Task (Step-by-Step Mode)

| Test Subtask | Test Model | Perception Score | Decision Score | Execution Steps | Composite Score | Normalized Composite Score |
|---|---|---|---|---|---|---|
| Search for the target vehicle | GPT-4o | 92.9 | 92.9 | 3.8 | 367.5 | 85.1 |
| | Claude-3.5 | 44.8 | 44.8 | 6.8 | 127.4 | 29.5 |
| | Gemini-1.5-Pro | 50.0 | 50.0 | 6.2 | 151.9 | 35.2 |
| | Qwen-VL-Max | 72.2 | 72.2 | 4.6 | 261.5 | 60.5 |
| Approach and observe the vehicle | GPT-4o | 90.4 | 90.4 | 5.0 | 412.6 | 81.0 |
| | Claude-3.5 | 70.0 | 70.0 | 6.0 | 302.4 | 59.4 |
| | Gemini-1.5-Pro | 70.0 | 70.0 | 6.0 | 302.4 | 59.4 |
| | Qwen-VL-Max | 84.0 | 84.0 | 5.0 | 383.4 | 75.3 |
| Identify the vehicle model | GPT-4o | 100.0 | 100.0 | 1.0 | 482.2 | 100.0 |
| | Claude-3.5 | 100.0 | 100.0 | 1.0 | 482.2 | 100.0 |
| | Gemini-1.5-Pro | 100.0 | 100.0 | 1.0 | 482.2 | 100.0 |
| | Qwen-VL-Max | 100.0 | 100.0 | 1.0 | 482.2 | 100.0 |

### 5.3.2 Dynamic Rider Tracking Task

**1) Task Description:** The UAV starts from a location with a clear but distant view of the target. Once the task begins, the target vehicle starts moving, and the UAV is required to maintain continuous tracking. During the process, the target passes through multiple intersections and executes several turns. This task is structured with a total of 14 distinct nodes, and the UAV must make accurate decisions at each task node to avoid losing the target. Due to the difficulty of decomposing the tracking process into distinct sub-stages, a "Step-by-Step" testing mode is not included.

**2) Evaluation Method:** This task focuses on evaluating the UAV-EA's perception and decision-making performance at each key task node. Following each run, we record the model's perception and decision scores at every turning point. Upon completion of the entire task, the average task score across all turning points is calculated. Each model is tested under identical conditions for five runs, and the final perception and decision scores are obtained by averaging the results from all runs. The overall composite score is computed as the average of the perception and decision scores.

**3) Results:** As shown in **Table 6**, Qwen-VL-Max demonstrated superior performance in the dynamic rider tracking task, achieving the highest composite score of 73.2% with notably balanced perception at 75.7% and decision capabilities at 70.7%. Gemini-1.5-Pro and GPT-4o delivered moderate performance with closely aligned composite scores of 57.9% and 57.2% respectively, though both exhibited a noticeable decline from perception to decision scores, indicating a shared weakness in converting visual understanding into effective control. Claude-3.5 trailed behind with the lowest composite score of 50.4%, reflecting consistent challenges in both perceiving the target and making subsequent tracking decisions.

The evaluation revealed two primary limitations common to most models in this dynamic tracking scenario:

Firstly, the models demonstrated inadequate understanding of motion continuity and behavioral prediction. This was particularly evident at intersections where the target executed turns. Claude-3.5's low scores of 52.9% in perception and 47.9% in decision highlight a fundamental failure in anticipating rider movement, often resulting in complete target loss. While GPT-4o and Gemini-1.5-Pro achieved reasonable perception scores of 61.4%, their significant drop in decision scores to 52.9% and 54.3% respectively underscores a systematic inability to translate perceptual observations into anticipatory control commands.

Secondly, all models exhibited deficiencies in distinguishing target-related cues from environmental distractions and leveraging historical information for perception and decision-making. As illustrated in Fig. 12, when confronted with a sudden target disappearance scenario, none of the models could identify the correct situation where the target was concealed in a roadside parking area. Instead, they were misled by various distractors including tree shadows, intersection turns, and camera blind spots. This consistent failure demonstrates the models' limited capacity to effectively utilize historical trajectory information and accurately differentiate genuine target cues from environmental noise, ultimately leading to erroneous decisions in dynamic tracking contexts. The performance degradation observed across all models, particularly in decision-making, points to a generalized weakness in maintaining a stable spatial reference frame and executing coherent tracking strategies through complex, changing scenarios.

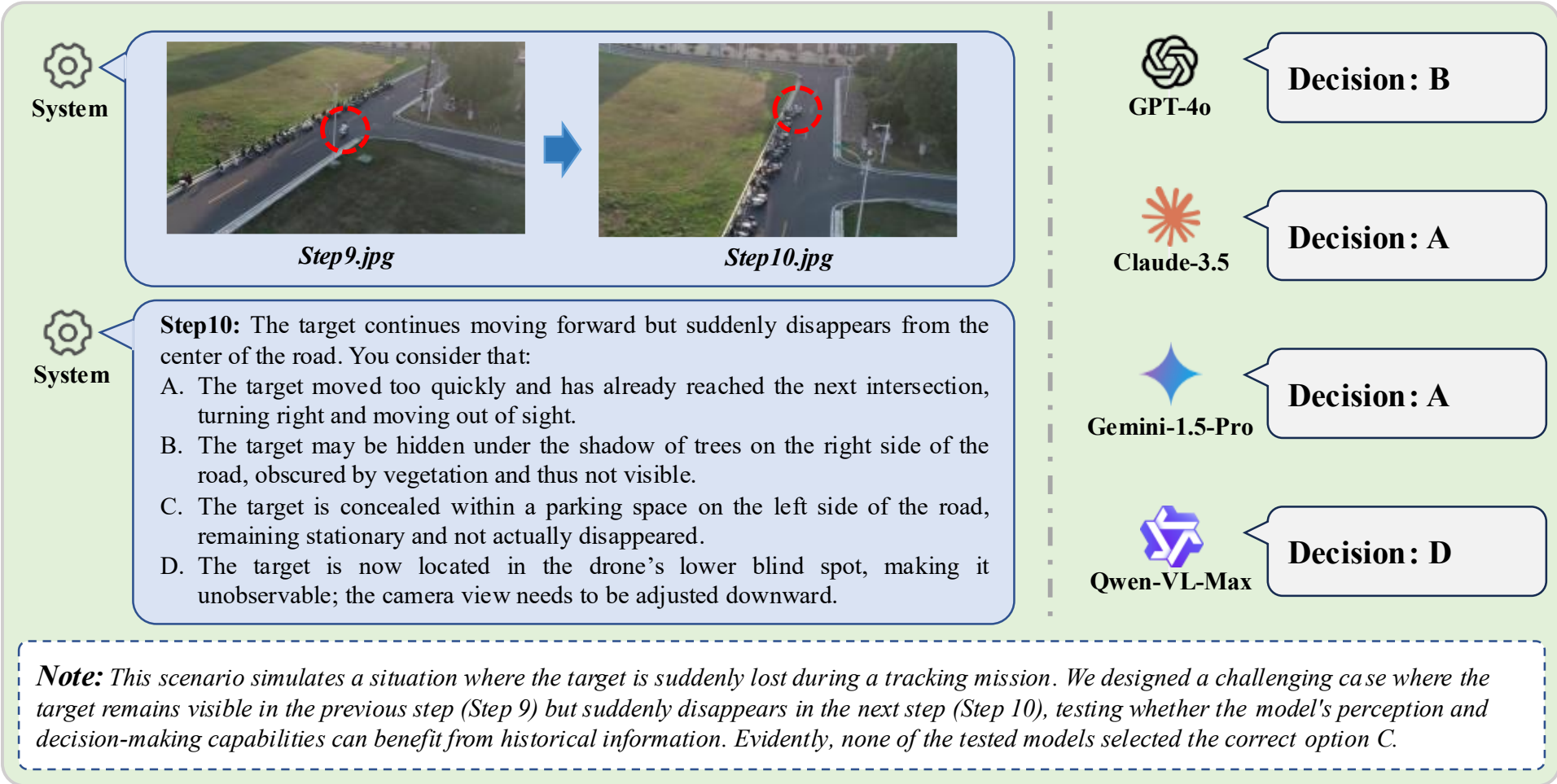

Fig. 12 Example of the model responses in the dynamic rider tracking task. The red circles in both figures are provided for illustrative purposes to help readers focus on the task objectives; they were not present in the images used during the model testing process.

Table 6: Experimental Results of the Dynamic Rider Tracking Task

| Test Model | Perception Score | Decision Score | Composite Score |
|---|---|---|---|
| GPT-4o | 61.4 | 52.9 | 57.2 |
| Claude-3.5 | <u>**52.9**</u> | <u>**47.9**</u> | <u>**50.4**</u> |
| Gemini-1.5-Pro | 61.4 | 54.3 | 57.9 |
| Qwen-VL-Max | **75.7** | **70.7** | **73.2** |

### 5.4 Experimental Results in the Dynamic-Virtual Test Environment

#### 5.4.1 Cargo Delivery Task

**1) Task Description:** The UAV is initialized at a distant starting point away from the designated seaport target. Given the spatial coordinates of a seaport, the UAV-EA is tasked with navigating to the port and delivering onboard cargo to a specific target vessel docked there. The task is structured into three sequential phases: "Navigate to the port," "Search for the cargo vessel," and "Approach the target vessel." The shortest possible execution path requires 3 steps, distributed across the three phases as 1, 1, and 1 step, respectively.

**2) Evaluation Method:** The evaluation procedure follows the same methodology as in the Fine-Grained Vehicle Recognition Task. In the end-to-end setting, each model is tested five times under identical conditions, and the average results are reported in Table 7. In the step-by-step setting, each subtask is evaluated over four repetitions, with results summarized in Table 8. During metric computation, we set the composite scoring parameters as follows: $\alpha = 1.1$, $b = 0.5$. For the end-to-end task, the step threshold is set to $Step' = 25$. For the step-by-step setting, the thresholds are defined as $Step' = 5$ for "Navigate to the port," $Step' = 10$ for "Search for the cargo vessel," and $Step' = 20$ for "Approach the target vessel."

**3) Results:** As shown in Table 7, the end-to-end testing results reveal that among the evaluated

VLMs, GPT-4o achieved the highest score of 60.2%, while Qwen-VL-Max scored only 8.8%. These results emphasize a major deficiency in autonomous task decomposition and multi-step coordination. For instance, GPT-4o achieved 93.3% in perception accuracy, but its decision score of 84.1% and execution steps of 9.8 suggest inefficiencies in dynamic path planning. The theoretical optimal step count is around 5, but actual performance nearly doubled due to the model's inability to balance conflicting goals including minimizing distance while avoiding obstacles and adapting strategies in real time. Excessive trajectory corrections resulted in redundant movements and increased step counts. Claude-3.5 and Gemini-1.5-Pro further illustrate this issue, with decision scores of 33.0% and 62.3% respectively. Claude-3.5 often falls into local optima during dynamic obstacle avoidance, while Gemini-1.5-Pro suffers from fragmented path planning caused by breakdowns in logical reasoning.

In the step-by-step testing results from **Table 8**, performance fragmentation among models is even more apparent across sub-tasks. In the simplest task, Navigate to the port, all models performed perfectly with 100% decision accuracy. However, performance sharply dropped in more complex stages. For example, in the Approach the target vessel stage, Qwen-VL-Max's decision score plummeted to 27.6%, execution steps increased to 18.3, and its composite score dropped to 20.6%. This indicates deeper issues in multi-modal reasoning. Although the model could locate the vessel visually with a 79.0% perception score, it failed to integrate spatial coordinates with dynamic environmental factors such as wind speed or vessel motion, resulting in poor decisions. It continued following an outdated path based on the vessel's initial coordinates, ignoring its real-time movement, and thus executed ineffective actions. The perception-decision gaps for Claude-3.5 at 78.5% versus 40.2% and Gemini-1.5-Pro at 57.4% versus 47.3% further highlight the inability of current models to effectively integrate visual inputs with spatial reasoning. Their decisions often rely on superficial semantic cues rather than physics-grounded inference. Additionally, Qwen-VL-Max's poor performance in the Search for vessel task with 6.8 steps and a 33.5% score reveals its limited adaptability to dynamic conditions. It failed to adjust its strategy in response to real-time updates such as vessel movement, rigidly following preplanned sequences.

Table 7: Experimental Results of the Cargo Delivery Task in Virtual Environment (End-to-End Mode)

| Test Model | Perception Score ↑ | Decision Score ↑ | Execution Steps ↓ | Composite Score ↑ | Normalized Composite Score ↑ |
|---|---|---|---|---|---|
| GPT-4o | **93.3** | **84.1** | **9.8** | **346.3** | **60.2** |
| Claude-3.5 | 74.1 | 33.0 | 20.6 | 130.0 | 22.6 |
| Gemini-1.5-Pro | **<u>67.9</u>** | 62.3 | 15.6 | 196.9 | 34.2 |
| Qwen-VL-Max | 70.0 | **<u>30.8</u>** | **<u>25.0</u>** | **<u>50.4</u>** | **<u>8.8</u>** |

Synthesizing the results from both evaluation modes, the core limitations of current models can be categorized into three areas:

Firstly, current models demonstrate a notable lack of autonomous task decomposition capability. In end-to-end testing, execution steps consistently exceeded optimal values, as evidenced by GPT-4o requiring 9.8 steps compared to the theoretical minimum of approximately 5. This indicates models struggle to generate coherent sub-goals, particularly when encountering dynamic obstacles. The underlying issue stems from their dependence on static knowledge bases and fixed rule sets, which substantially limits their adaptability to environmental changes. As illustrated in **Fig. 13**, certain models exhibit logical inconsistencies between perception results and subsequent decisions. For instance, even after correctly identifying the cargo vessel, models may still command the UAV to continue searching rather than initiating approach maneuvers.

Secondly, multimodal reasoning capabilities remain underdeveloped across all evaluated models. Step-by-step testing revealed substantial disparities between perception and decision scores, with Gemini-1.5-Pro showing a 10.1 percentage point gap between its 57.4% perception score and 47.3% decision score. These results highlight fundamental deficiencies in cross-modal information integration. While models can typically identify visual features of targets such as cargo vessels, they frequently fail to spatially ground this information within appropriate coordinate systems. This limitation directly impairs path planning effectiveness and creates disconnects between perceptual understanding and decision outputs.

Thirdly, domain generalization presents a significant challenge for current systems. Qwen-VL-Max performed consistently poorly across all testing modes. It achieved only 8.8% in the end-to-end evaluation, and 33.5% and 20.6% in the step-by-step settings. This indicates that even large, closed-source models trained on extensive datasets remain heavily dependent on domain-specific training for specialized tasks like dynamic port operations. General-purpose pretraining appears insufficient for capturing crucial domain knowledge elements including UAV kinematics and spatial schemas, ultimately leading to nearly random performance in real-world applications.

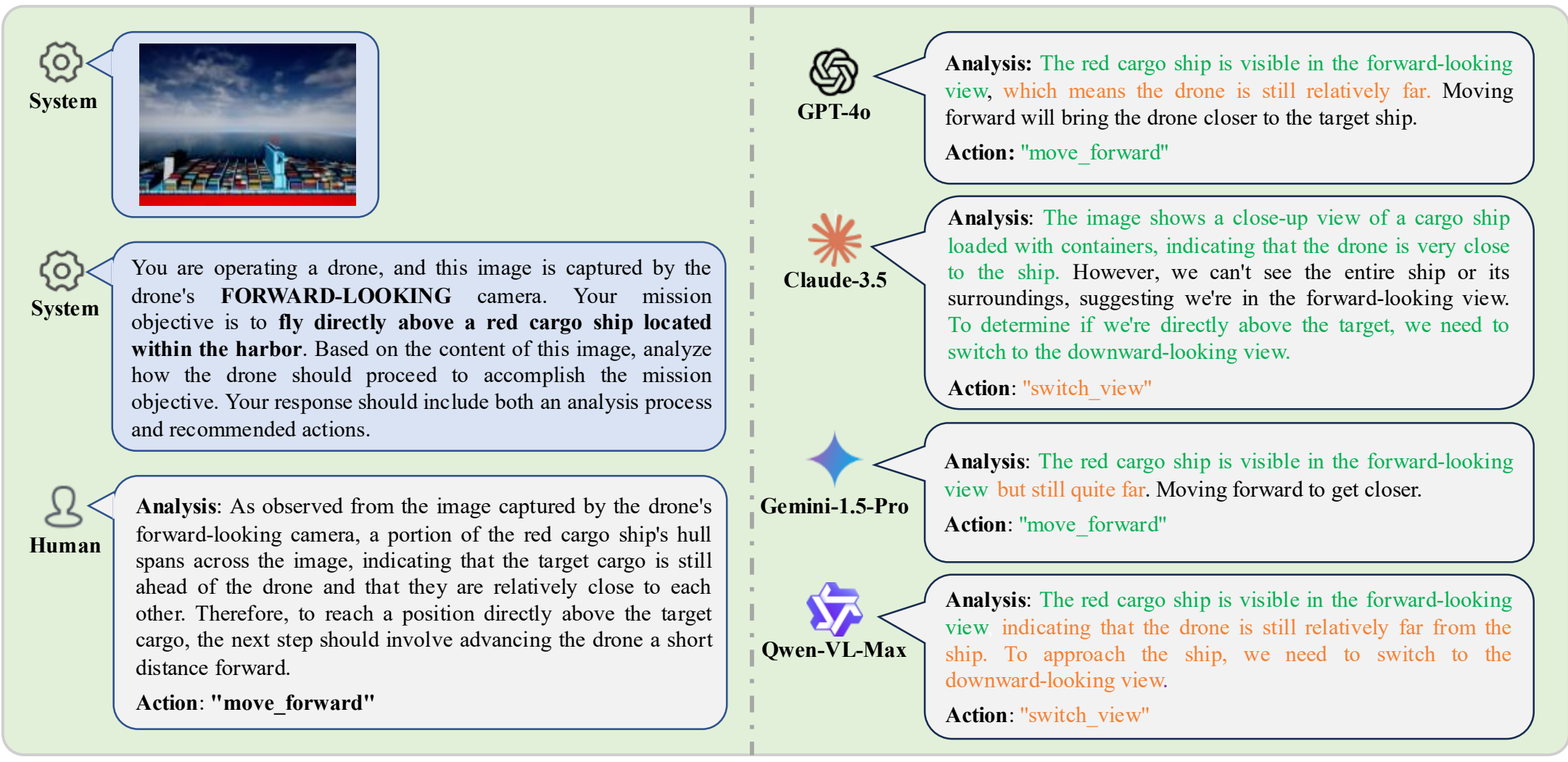

Fig. 13 Example of the model responses in the cargo delivery task. The left shows the system prompt and human reference; the right shows the model's analysis and decision. Green indicates correct responses, orange indicates errors.

Table 8: Experimental Results of the Cargo Delivery Task in Virtual Environment (Step-by-Step Mode)

| Test Subtask | Test Model | Perception Score | Decision Score | Execution Steps | Composite Score | Normalized Composite Score |
|---|---|---|---|---|---|---|
| Navigate to the port | GPT-4o | \ | 100.0 | 1.0 | 482.2 | 100.0 |
| | Claude-3.5 | \ | 100.0 | 1.0 | 482.2 | 100.0 |
| | Gemini-1.5-Pro | \ | 100.0 | 1.0 | 482.2 | 100.0 |
| | Qwen-VL-Max | \ | 100.0 | 1.0 | 482.2 | 100.0 |
| Search for the cargo vessel | GPT-4o | 100.0 | 100.0 | 2.0 | 482.2 | 89.6 |
| | Claude-3.5 | 100.0 | 100.0 | 2.0 | 482.2 | 89.6 |
| | Gemini-1.5-Pro | 100.0 | 100.0 | 2.0 | 482.2 | 89.6 |
| | Qwen-VL-Max | 100.0 | 26.9 | 6.8 | 180.4 | 33.5 |
| Approach the target vessel | GPT-4o | 100.0 | 96.9 | 7.3 | 395.9 | 69.6 |
| | Claude-3.5 | 78.5 | 40.2 | 16.0 | 147.9 | 26.0 |
| | Gemini-1.5-Pro | 57.4 | 47.3 | 14.0 | 145.6 | 25.6 |
| | Qwen-VL-Max | 79.0 | 27.6 | 18.3 | 117.0 | 20.6 |

### 5.4.2 Building Firefighting Task

**1) Task Description:** The UAV-EA is initially positioned at a distant starting location relative to the burning building. Given the spatial coordinates of the target building, the agent is required to navigate to the site and perform a firefighting operation. Specifically, the agent must first reach the fire-exposed side of the building, then accurately locate the fire source, and finally extinguish it using the onboard firefighting equipment. The task is divided into three sequential stages: "Navigate to the fire scene," "Locate the fire source," and "Execute firefighting operation." The shortest possible execution path requires 3 steps, distributed across the three phases as 1, 1, and 1 step, respectively.

**2) Evaluation Method:** The evaluation procedure follows the same methodology as in the Fine-Grained Vehicle Recognition Task. In the end-to-end setting, each model is tested five times under identical conditions, and the average results are reported in Table 9. In the step-by-step setting, each subtask is evaluated over four repetitions, with results summarized in Table 10. During metric computation, we set the composite scoring parameters as follows: $\alpha = 1.1$, $b = 0.5$. For the end-to-end task, the step threshold is set to $Step' = 25$. For the step-by-step setting, the thresholds are defined as $Step' = 5$ for "Navigate to the fire scene," $Step' = 10$ for "Search for the fire source," and $Step' = 10$ for "Execute firefighting operation."

**3) Results:** In the end-to-end evaluation presented in Table 9, GPT-4o achieved the highest normalized composite score of 57.2% but required 10.2 execution steps, substantially exceeding the theoretical optimum of approximately 6 steps. This indicates limited adaptability to dynamic environmental conditions, as the model's inability to respond promptly to fire propagation resulted in repeated path corrections. Other models demonstrated more severe limitations: Claude-3.5 and Qwen-VL-Max attained scores of only 12.3% and 11.2% respectively, while Gemini-1.5-Pro scored merely 0.3%. Gemini-1.5-Pro's near-zero perception and decision accuracy at 0.0% and 4.0% respectively, combined with its requirement of 25 steps, reveals fundamental failures in both autonomous task decomposition and multimodal reasoning. As illustrated in Fig. 14, even in straightforward target localization scenarios where other models succeeded, Gemini-1.5-Pro consistently generated incorrect decisions.

The step-by-step evaluation results in Table 10 revealed pronounced task dependency across all models. All systems performed flawlessly on simpler sub-tasks including Navigate to the fire scene and Search for the fire source, achieving perfect scores in perception and decision metrics. However, performance deteriorated significantly during the critical Execute firefighting operation sub-task. GPT-4o attained 81.7% in perception and 76.7% in decision while requiring 5 steps, yet demonstrated difficulties in tactical planning such as spray angle adjustment and suppressant volume control. Claude-3.5 achieved perfect scores in sub-task evaluation but its end-to-end performance of only 12.3% revealed insufficient temporal coordination. Similarly, Qwen-VL-Max scored 61.2% in firefighting sub-tasks but achieved only 11.2% in integrated evaluation, indicating poor cross-phase integration. Gemini-1.5-Pro maintained inadequate performance with 55.0% perception and 45.0% decision scores, requiring 10 steps and achieving a composite score of merely 9.3%.

Table 9: Experimental Results for Building Firefighting Task in Virtual Scenario (End-to-End)

| Test Model | Perception Score ↑ | Decision Score ↑ | Execution Steps ↓ | Composite Score ↑ | Normalized Composite Score ↑ |
|---|---|---|---|---|---|
| GPT-4o | **85.0** | **86.5** | **10.2** | **328.9** | **57.2** |
| Claude-3.5 | 39.1 | 27.4 | 23.6 | 70.7 | 12.3 |
| Gemini-1.5-Pro | <u>**0.0**</u> | <u>**4.0**</u> | <u>**25.0**</u> | <u>**2.0**</u> | <u>**0.3**</u> |
| Qwen-VL-Max | 65.0 | 64.0 | <u>**25.0**</u> | 64.5 | 11.2 |

Three fundamental limitations emerge from detailed analysis of the experimental results:

Firstly, models exhibit consistently inflexible planning in dynamic environments. GPT-4o's 10.2 execution steps substantially exceeded the 3-step theoretical optimum, demonstrating rigid planning strategies. The model's repeated path corrections and redundant maneuvers when confronting fire spread reveal inadequate real-time adjustment capabilities. This planning inflexibility directly contributes to inefficient task execution and excessive resource consumption across all tested models.

Secondly, ineffective multimodal integration severely limits tactical decision-making. While GPT-4o achieved 81.7% in perception accuracy during firefighting operations, it struggled with spray angle adjustment and suppressant volume control. Gemini-1.5-Pro's near-zero accuracy in end-to-end tasks further exemplifies the failure to integrate visual cues with domain knowledge. This integration gap prevents models from transforming perceptual information into effective operational strategies.

Thirdly, weak coordination across task stages causes significant performance fragmentation. Claude-3.5's perfect sub-task scores contrasted sharply with its 12.3% end-to-end performance, highlighting critical coordination deficiencies. Qwen-VL-Max's 61.2% firefighting sub-task score dropping to 11.2% in full task execution demonstrates the cascading effect of stage transition failures. This coordination weakness undermines the models' ability to maintain consistent performance throughout multi-stage operations.

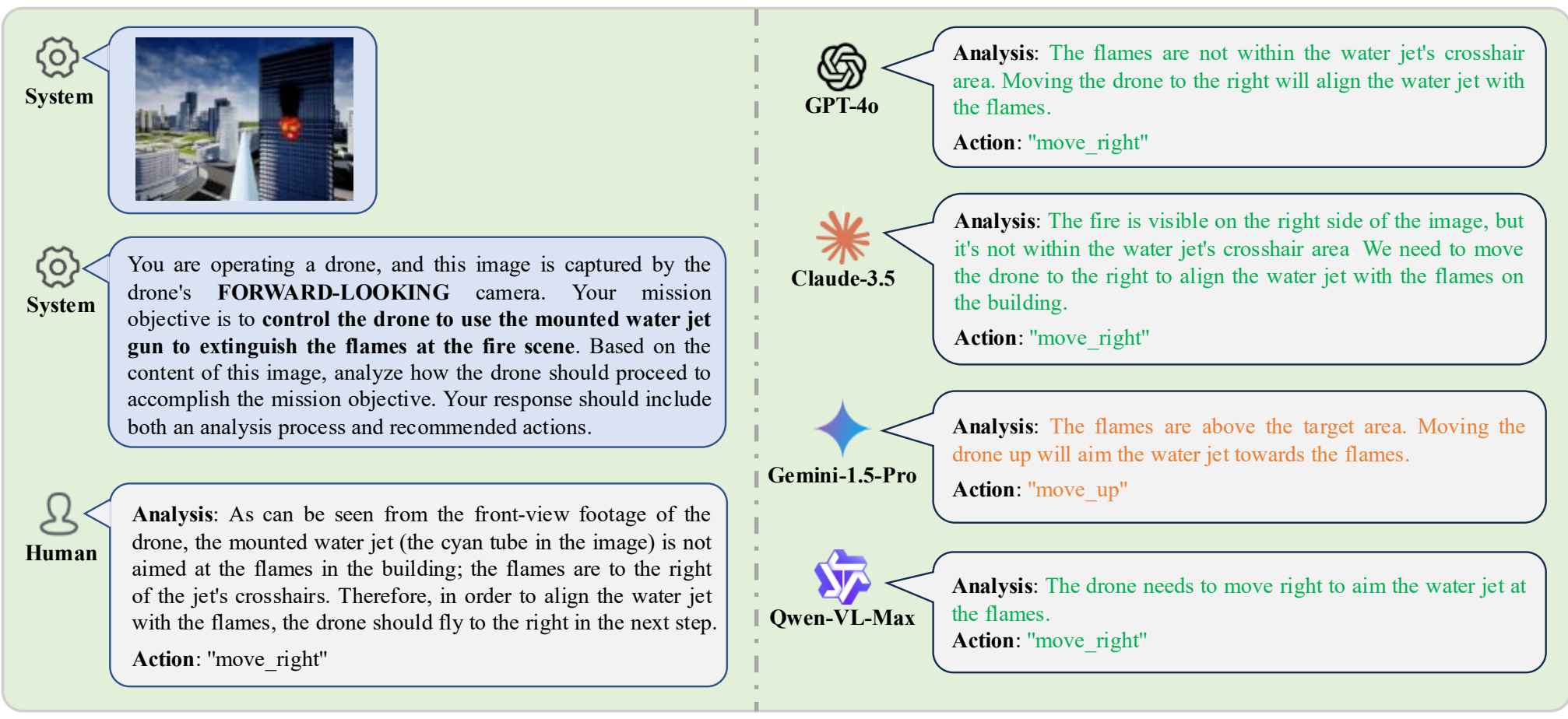

Fig. 14 Example of the model responses in the building firefighting task. The left shows the system prompt and human reference; the right shows the model's analysis and decision. Green indicates correct responses, orange indicates errors.

Table 10: Experimental Results for Building Firefighting Task in Virtual Scenario (Step-by-Step)

| Test Subtask | Test Model | Perception Score | Decision Score | Execution Steps | Composite Score | Normalized Composite Score |
|---|---|---|---|---|---|---|
| Navigate to the fire scene | GPT-4o | \ | 100.0 | 1.0 | 482.2 | 100.0 |
| | Claude-3.5 | \ | 100.0 | 1.0 | 482.2 | 100.0 |
| | Gemini-1.5-Pro | \ | 100.0 | 1.0 | 482.2 | 100.0 |
| | Qwen-VL-Max | \ | 100.0 | 1.0 | 482.2 | 100.0 |
| Locate the fire source | GPT-4o | 100.0 | 100.0 | 3.0 | 432.0 | 80.3 |
| | Claude-3.5 | 100.0 | 100.0 | 3.0 | 432.0 | 80.3 |
| | Gemini-1.5-Pro | 100.0 | 100.0 | 3.0 | 432.0 | 80.3 |
| | Qwen-VL-Max | 100.0 | 100.0 | 3.0 | 432.0 | 80.3 |
| Execute firefighting operation | GPT-4o | 81.7 | 76.7 | 5.0 | 274.5 | 51.1 |
| | Claude-3.5 | 100.0 | 100.0 | 4.0 | 387.0 | 71.9 |
| | Gemini-1.5-Pro | 55.0 | 45.0 | 10.0 | 50.0 | 9.3 |
| | Qwen-VL-Max | 90.0 | 90.0 | 4.5 | 329.6 | 61.2 |

### 5.4.3 Moving Target Tracking Task

**1) Task Description:** This task is similar to the Dynamic Rider Tracking Task, the UAV begins its operation from a position where the target is clearly visible but at a considerable distance. Throughout the scenario, the target proceeds through multiple intersections and executes a series of turns. The trajectory of the target vehicle, including the number and locations of all turns, remains fixed for all task instances. Due to the inherent difficulty in decomposing the tracking process into distinct subtasks, this task does not include a "step-by-step" testing mode.

**2) Evaluation Method:** Similar to the Dynamic Rider Tracking Task, this task focuses specifically on the UAV-EA's performance at key behavioral transition points—namely, when the target vehicle turns at intersections. After each run, we evaluate the model's perception and decision performance at each turn, then compute the average task score across all turning points. Each model is tested five times under identical conditions, and the final perception and decision scores are obtained by averaging across all runs. The overall composite score is calculated as the mean of the perception and decision scores.

**3) Results:** As shown in Table 11, all evaluated models demonstrated suboptimal performance in the moving target tracking task, with none achieving a composite score above 40%, highlighting the considerable challenges posed by dynamic spatial reasoning. Claude-3.5 achieved the best overall performance with a composite score of 39.5%, including 42.1% in perception and 36.8% in decision-making. Despite outperforming other models, Claude-3.5 still struggled to accurately respond to behavioral transitions of the target, such as detecting turning intent at intersections, often resulting in trajectory deviation or delayed responses.

GPT-4o and Gemini-1.5-Pro scored 21.1% and 26.3%, respectively, reflecting their limitations in handling dynamic input streams. GPT-4o exhibited notable weaknesses in perception, suggesting difficulty in interpreting visual sequences from UAV viewpoints. In contrast, Gemini-1.5-Pro showed relatively balanced but low perception and decision scores, indicating difficulty in transforming perceptual observations into coherent tracking decisions under motion variability. Qwen-VL-Max performed the worst across all dimensions, with both perception and decision scores at 15.8%, representing a near-complete failure in understanding target movement. The model frequently misidentified direction changes or failed to issue any meaningful response, suggesting a breakdown in motion recognition and adaptive control.

Table 11: Experimental Results of the Moving Target Tracking Task

| Test Model | Perception Score | Decision Score | Composite Score |
| --- | --- | --- | --- |
| GPT-4o | 21.1 | 21.1 | 21.1 |
| Claude-3.5 | **42.1** | **36.8** | **39.5** |
| Gemini-1.5-Pro | 26.3 | 26.3 | 26.3 |
| Qwen-VL-Max | **15.8** | **15.8** | **15.8** |

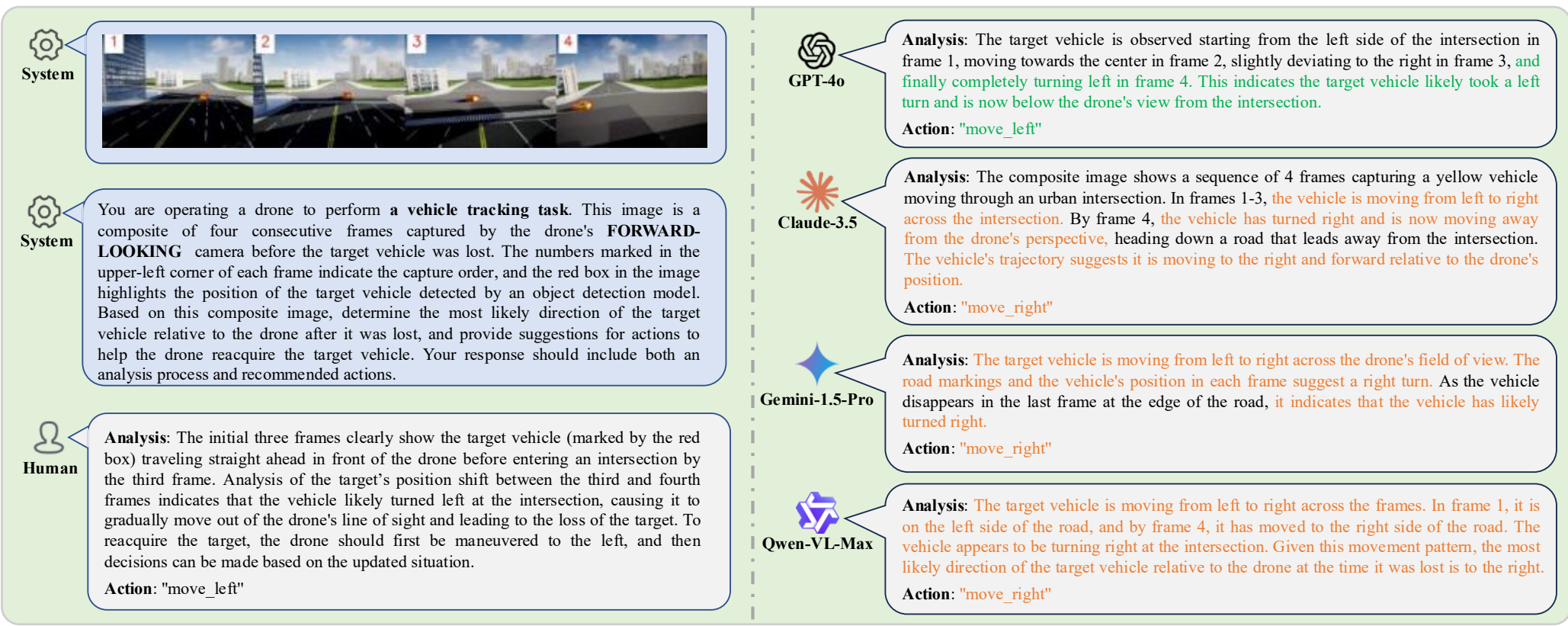

Fig. 15 Example of the model responses in the moving target tracking task. The left shows the system prompt and human reference; the right shows the model's analysis and decision. Green indicates correct responses, orange indicates errors.

During testing, two major deficiencies were consistently observed across all evaluated models:

Firstly, the models demonstrated significant limitations in interpreting egocentric visual streams from the UAV's operational perspective. As illustrated in Fig. 15, changes in positional relationships, viewing angles, and lighting conditions introduced substantial visual noise that disrupted perception accuracy. The models predominantly relied on static image coordinate systems rather than dynamic spatial reasoning, resulting in frequent misjudgments of turning directions and movement trajectories. This fundamental weakness in processing first-person visual sequences directly contributed to erroneous tracking decisions at critical navigation points.

Secondly, all models exhibited a critical lack of consistent spatial reference maintenance during tracking operations. Without robust mechanisms to preserve spatial coherence across temporal sequences, the systems failed to continuously update their environmental understanding based on frame-to-frame positional and orientational changes. This deficiency led to fragmented tracking strategies that could not adapt to the target's dynamic movement patterns. For instance, several models maintained inappropriate UAV postures despite clear evidence of target trajectory changes, ultimately resulting in complete target loss due to this spatial reasoning breakdown.

## 6. Discussion

In this paper, we propose BEDI, a systematic and standardized benchmark designed for UAV-EAs, designed to address key limitations in existing benchmarks such as fragmented task definitions, lack of environmental dynamics, and insufficient platform openness. By introducing the Dynamic Chain-of-Embodied-Task paradigm, we decompose complex UAV-EA tasks into multi-stage perception–decision–action loops and define a comprehensive metric system that assesses six core capabilities: semantic perception, spatial perception, motion control, tool utilization, task planning, and action generation. This approach enables, for the first time, fine-grained and end-to-end quantification of UAV-EA performance. Furthermore, we develop a hybrid and open testing platform that incorporates a wide range of both virtual and real-world scenarios. Experimental results on BEDI reveal significant limitations in current VLMs when applied to UAV-EA tasks, particularly in dynamic decision-making and spatial reasoning. Additional challenges include poor generalization and difficulties in understanding first-person perspectives, underscoring critical areas for future improvement in UAV-EA systems.

While BEDI addresses several key gaps in UAV-EA evaluation, it also has certain limitations. The current framework emphasizes perception, decision-making, and action execution but does not incorporate higher-order cognitive abilities such as memory and prediction, which are essential for long-term autonomous operations. Furthermore, the simulation environment does not account for physical disturbances like weather variations or sensor noise, potentially overestimating model robustness in real-world conditions. The evaluation pipeline also relies exclusively on 2D visual inputs, without incorporating 3D sensory modalities such as LiDAR or depth maps, which limits alignment with practical UAV applications. Thus, future work will focus on three main directions to enhance BEDI's capabilities: first, integrating memory and prediction mechanisms to enable the assessment of long-term reasoning and foresight; second, introducing realistic physical disturbances to improve simulation fidelity and model robustness; and third, expanding the input space to include 3D sensory data such as LiDAR and depth maps, in order to better reflect real-world UAV perception requirements.

## Acknowledgements

This work was supported in part by the Natural Science Foundation of Hunan for Distinguished Young Scholars under Grant 2022JJ10072; in part by the National Natural Science Foundation of China under Grant 42471419 and Grant 42171376.